\pdfoutput=1
\documentclass{article}
\usepackage{microtype}
\usepackage{booktabs,amsmath,amssymb,graphicx,xcolor,url,multirow}
\usepackage{array}
\usepackage{pgfplots}\pgfplotsset{compat=1.18}\usetikzlibrary{patterns}
\usepackage[colorlinks=true,linkcolor=blue,citecolor=blue,urlcolor=blue]{hyperref}
\usepackage{xurl}
\usepackage{adjustbox}
\usepackage{placeins}
\AtBeginDocument{\hypersetup{pdfsubject={Preprint}}}
\usepackage[preprint]{icml2026}

\newcommand{\finB}{fin@1$(B)$}
\newcommand{\relB}{pass@1$(B\,|\,\text{fin})$}

\newcommand{\armF}{arm~F}
\newcommand{\armH}{arm~H}
\newcommand{\armI}{arm~I}
\newcommand{\armJ}{arm~J}
\newcommand{\armP}{arm~P}
\newcommand{\armK}{arm~K}
\newcommand{\armM}{arm~M}
\newcommand{\armO}{arm~O}
\newcommand{\armB}{arm~B16}
\newcommand{\armR}{arm~R}
\newcommand{\armRW}{arm~RW}
\newcommand{\armD}{arm~D16}
\newcommand{\armS}{arm~S}

\icmltitlerunning{What Outcome RL Buys: Finishing, Reliability, Rarely Reach}

\DeclareRobustCommand{\textbf}[1]{{\bfseries\boldmath #1}}
\usepackage{tabularx}
\begin{document}

\twocolumn[
\icmltitle{What Outcome RL Buys: Finishing, Reliability, Rarely Reach}
\begin{icmlauthorlist}
\icmlauthor{Shan Yang}{ind}
\end{icmlauthorlist}
\icmlaffiliation{ind}{Individual Researcher}
\icmlcorrespondingauthor{Shan Yang}{alexyangshan@gmail.com}
\icmlkeywords{reinforcement learning, reasoning, language models, physics}
\vskip 0.3in
]
\printAffiliationsAndNotice{}
\makeatletter\gdef\@icmltitlerunning{What Outcome RL Buys: Finishing, Reliability, Rarely Reach}\makeatother

\begin{abstract}
Outcome RL raises pass@1 but not pass@$k$ at large $k$ \citep{yue2025}. But every pass@1 is scored under a token budget. Every recipe also trains under one. On physics and math problems at fixed token budgets, we compare the pass@1 of $20$ released RL models ($1.5$B to $32$B) with that of the thinking models they started from. We ask whether each pass@1 gain is changed behaviour or new knowledge. We divide each change in pass@1 into three parts. \emph{Finishing} measures answering before the budget runs out. \emph{Reliability} measures being right once finished. \emph{Reach} measures problems RL solves that its base answered wrongly in all eight tries, minus problems the base solves that RL answered wrongly in all eight. We find four things. (1) Where the base often hits the token budget before answering, most of the change, gain or loss, is finishing ($82\%$ at $4$k and $8$k). Given more tokens, the base closes most of the finishing gap itself. (2) Where the base has enough tokens, the gain is reliability: the RL model is graded right more often, even on problems both finish. (3) By our definition, RL rarely solves what its base cannot. Reach adds at most $2.1$ percentage points on any of our four physics and math sets. We checked $144$ physics problems that only the RL model solved. Of these, $3.5\%$ hold up when resampled or read. In a random sample, nine of ten new RL steps a human confirmed also appear in the base's samples. (4) In a small-scale RL run on an $8$B model, the gain on new problems is finishing, with reliability flat. Supplying each problem's physical setup was not shown to improve on plain RL.
\end{abstract}
\noindent Project page: \url{https://shanyang.me/what-outcome-rl-buys/}. Code and released tables: \url{https://github.com/shanyang-me/outcome-rl-buys}.

\section{Introduction}
\label{sec:intro}
\begin{figure}[!t]
\centering
\definecolor{finishink}{RGB}{16,50,88}
\definecolor{reliaink}{RGB}{242,146,24}
\definecolor{reachink}{RGB}{0,158,115}
\resizebox{\linewidth}{!}{\begin{tikzpicture}[x=24.0cm,y=0.9cm]
\node[anchor=east,font=\footnotesize\itshape,text=black!70,inner sep=1pt] at (-0.0640,0.36) {at the training cap, physics};
\fill[finishink] (0.00000,-0.19) rectangle (0.07793,0.19);
\fill[reliaink] (0.07793,-0.19) rectangle (0.07948,0.19);
\fill[pattern=north east lines,pattern color=reachink] (-0.00250,-0.19) rectangle (0.00000,0.19);
\node[anchor=west,font=\footnotesize,inner sep=1pt] at (0.08248,0.00) {$+0.077$};
\node[anchor=east,font=\footnotesize,inner sep=1pt] at (-0.0640,0.00) {base finishes $<90\%$ ($4$)};
\fill[finishink] (0.00000,-0.65) rectangle (0.00155,-0.27);
\fill[reliaink] (0.00155,-0.65) rectangle (0.03359,-0.27);
\fill[pattern=north east lines,pattern color=reachink] (0.03359,-0.65) rectangle (0.03823,-0.27);
\node[anchor=west,font=\footnotesize,inner sep=1pt] at (0.04123,-0.46) {$+0.038$};
\node[anchor=east,font=\footnotesize,inner sep=1pt] at (-0.0640,-0.46) {base finishes $\geq90\%$ ($16$)};
\fill[finishink] (0.00000,-1.11) rectangle (0.01683,-0.73);
\fill[reliaink] (0.01683,-1.11) rectangle (0.04277,-0.73);
\fill[pattern=north east lines,pattern color=reachink] (0.04277,-1.11) rectangle (0.04598,-0.73);
\node[anchor=west,font=\footnotesize,inner sep=1pt] at (0.04898,-0.92) {$+0.046$};
\node[anchor=east,font=\footnotesize,inner sep=1pt] at (-0.0640,-0.92) {all ($20$)};
\node[anchor=east,font=\footnotesize\itshape,text=black!70,inner sep=1pt] at (-0.0640,-1.42) {next budget above the cap, physics};
\fill[finishink] (0.00000,-1.97) rectangle (0.02869,-1.59);
\fill[reliaink] (0.02869,-1.97) rectangle (0.05573,-1.59);
\fill[pattern=north east lines,pattern color=reachink] (0.05573,-1.97) rectangle (0.05771,-1.59);
\node[anchor=west,font=\footnotesize,inner sep=1pt] at (0.06071,-1.78) {$+0.058$};
\node[anchor=east,font=\footnotesize,inner sep=1pt] at (-0.0640,-1.78) {base finishes $<90\%$ ($4$)};
\fill[finishink] (0.00000,-2.43) rectangle (0.00285,-2.05);
\fill[reliaink] (0.00285,-2.43) rectangle (0.03393,-2.05);
\fill[pattern=north east lines,pattern color=reachink] (0.03393,-2.43) rectangle (0.03927,-2.05);
\node[anchor=west,font=\footnotesize,inner sep=1pt] at (0.04227,-2.24) {$+0.039$};
\node[anchor=east,font=\footnotesize,inner sep=1pt] at (-0.0640,-2.24) {base finishes $\geq90\%$ ($16$)};
\fill[finishink] (0.00000,-2.89) rectangle (0.00801,-2.51);
\fill[reliaink] (0.00801,-2.89) rectangle (0.03829,-2.51);
\fill[pattern=north east lines,pattern color=reachink] (0.03829,-2.89) rectangle (0.04296,-2.51);
\node[anchor=west,font=\footnotesize,inner sep=1pt] at (0.04596,-2.70) {$+0.043$};
\node[anchor=east,font=\footnotesize,inner sep=1pt] at (-0.0640,-2.70) {all ($20$)};
\node[anchor=east,font=\footnotesize\itshape,text=black!70,inner sep=1pt] at (-0.0640,-3.20) {at the training cap, competition math};
\fill[finishink] (0.00000,-3.75) rectangle (0.06382,-3.37);
\fill[reliaink] (0.06382,-3.75) rectangle (0.07936,-3.37);
\fill[pattern=north east lines,pattern color=reachink] (0.07936,-3.75) rectangle (0.08227,-3.37);
\node[anchor=west,font=\footnotesize,inner sep=1pt] at (0.08527,-3.56) {$+0.082$};
\node[anchor=east,font=\footnotesize,inner sep=1pt] at (-0.0640,-3.56) {base finishes $<90\%$ ($11$)};
\fill[finishink] (-0.04967,-4.21) rectangle (0.00000,-3.83);
\fill[reliaink] (0.00000,-4.21) rectangle (0.11967,-3.83);
\fill[pattern=north east lines,pattern color=reachink] (0.11967,-4.21) rectangle (0.12889,-3.83);
\node[anchor=west,font=\footnotesize,inner sep=1pt] at (0.13189,-4.02) {$+0.079$};
\node[anchor=east,font=\footnotesize,inner sep=1pt] at (-0.0640,-4.02) {base finishes $\geq90\%$ ($9$)};
\fill[finishink] (0.00000,-4.67) rectangle (0.01275,-4.29);
\fill[reliaink] (0.01275,-4.67) rectangle (0.07515,-4.29);
\fill[pattern=north east lines,pattern color=reachink] (0.07515,-4.67) rectangle (0.08090,-4.29);
\node[anchor=west,font=\footnotesize,inner sep=1pt] at (0.08390,-4.48) {$+0.081$};
\node[anchor=east,font=\footnotesize,inner sep=1pt] at (-0.0640,-4.48) {all ($20$)};
\draw[black!60,thin] (0,-4.79) -- (0,0.62);
\draw[black!60,thin] (-0.060,-4.79) -- (0.140,-4.79);
\foreach \x/\l in {-0.04/$-0.04$,0/$0$,0.04/$0.04$,0.08/$0.08$,0.12/$0.12$} \node[anchor=north,font=\scriptsize] at (\x,-4.81) {\l};
\node[anchor=north,font=\footnotesize] at (0.04,-5.12) {change in pass@1 over the base, mean over recipes};
\node[anchor=north,font=\footnotesize,inner sep=1pt,align=left] at (0.0300,-5.51) {\tikz[x=1cm,y=1cm]\fill[finishink] (0,0) rectangle (0.25,0.18); finishing $F$\quad \tikz[x=1cm,y=1cm]\fill[reliaink] (0,0) rectangle (0.25,0.18); reliability $P$\quad \tikz[x=1cm,y=1cm]\fill[pattern=north east lines,pattern color=reachink] (0,0) rectangle (0.25,0.18); reach, net of its mirror};
\end{tikzpicture}}
\caption{\sloppy Outcome RL's pass@1 gain on physics and competition math, split into finishing, reliability and net reach. Each bar is the mean over the recipes in its group. Recipes are grouped by how often their base finishes at the training cap on that set, with their number in parentheses. Reliability is measured on each model's own finished derivations. A part left of zero lowers the score: the RL model hits the budget before answering more often (finishing), is right less often once finished (reliability), or solves fewer problems alone than the base does (reach).}
\label{fig:overview}
\end{figure}
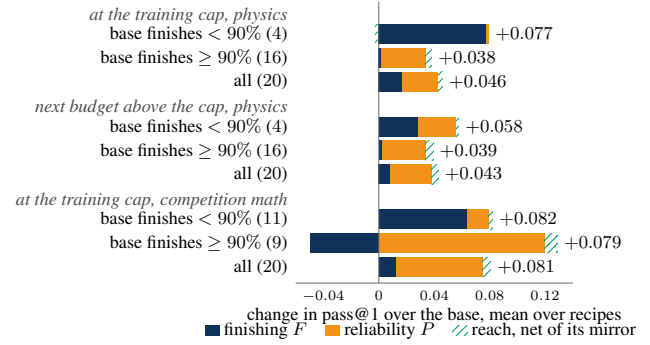
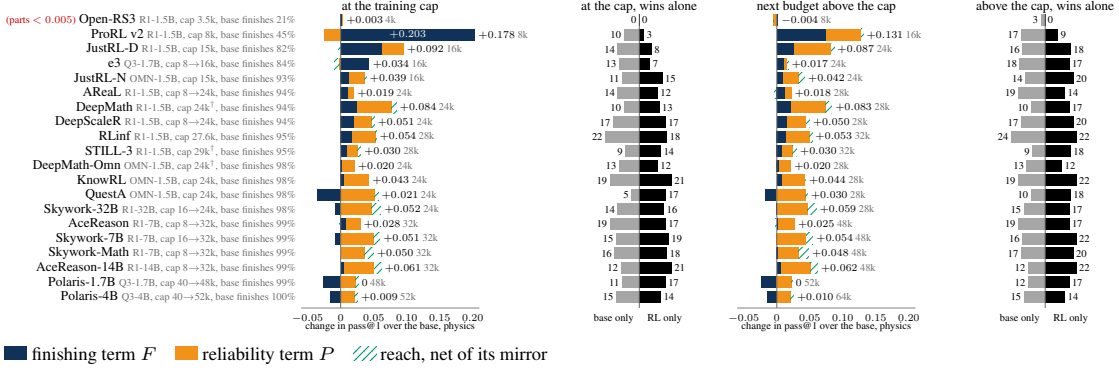
\begin{figure*}[!t]
\centering
\begin{minipage}{0.86\textwidth}
\definecolor{finishink}{RGB}{16,50,88}
\definecolor{reliaink}{RGB}{242,146,24}
\definecolor{reachink}{RGB}{0,158,115}
\resizebox{\textwidth}{!}{%
\begin{tikzpicture}[x=4.20cm,y=0.66cm,baseline=0pt]
\node[font=\small] at (0.5,0.47) {at the training cap};
\node[anchor=east,font=\footnotesize,inner sep=1pt] at (-0.02,0.00) {\textcolor{red!80!black}{\scriptsize (parts $<0.005$)} Open-RS3 \textcolor{black!55}{\scriptsize R1-1.5B, cap 3.5k, base finishes 21\%}};
\fill[finishink] (0.2222,-0.20) rectangle (0.2293,0.20);
\fill[reliaink] (0.2293,-0.20) rectangle (0.2346,0.20);
\node[anchor=west,font=\scriptsize,inner sep=1pt] at (0.2446,0.00) {$+0.003$ \textcolor{black!55}{4k}};
\node[anchor=east,font=\footnotesize,inner sep=1pt] at (-0.02,-0.52) {ProRL v2 \textcolor{black!55}{\scriptsize R1-1.5B, cap 8k, base finishes 45\%}};
\fill[finishink] (0.2222,-0.72) rectangle (0.9757,-0.32);
\node[font=\scriptsize,text=white,inner sep=0.5pt] at (0.5989,-0.52) {$+0.203$};
\fill[reliaink] (0.1277,-0.72) rectangle (0.2222,-0.32);
\node[anchor=west,font=\scriptsize,inner sep=1pt] at (0.9857,-0.52) {$+0.178$ \textcolor{black!55}{8k}};
\node[anchor=east,font=\footnotesize,inner sep=1pt] at (-0.02,-1.04) {JustRL-D \textcolor{black!55}{\scriptsize R1-1.5B, cap 15k, base finishes 82\%}};
\fill[finishink] (0.2222,-1.24) rectangle (0.4565,-0.84);
\fill[reliaink] (0.4565,-1.24) rectangle (0.5756,-0.84);
\fill[pattern=north east lines,pattern color=reachink] (0.2114,-1.24) rectangle (0.2222,-0.84);
\node[anchor=west,font=\scriptsize,inner sep=1pt] at (0.5856,-1.04) {$+0.092$ \textcolor{black!55}{16k}};
\node[anchor=east,font=\footnotesize,inner sep=1pt] at (-0.02,-1.56) {e3 \textcolor{black!55}{\scriptsize Q3-1.7B, cap 8$\to$16k, base finishes 84\%}};
\fill[finishink] (0.2222,-1.76) rectangle (0.3819,-1.36);
\fill[reliaink] (0.2153,-1.76) rectangle (0.2222,-1.36);
\fill[pattern=north east lines,pattern color=reachink] (0.1891,-1.76) rectangle (0.2153,-1.36);
\node[anchor=west,font=\scriptsize,inner sep=1pt] at (0.3919,-1.56) {$+0.034$ \textcolor{black!55}{16k}};
\node[anchor=east,font=\footnotesize,inner sep=1pt] at (-0.02,-2.08) {JustRL-N \textcolor{black!55}{\scriptsize OMN-1.5B, cap 15k, base finishes 93\%}};
\fill[finishink] (0.2222,-2.28) rectangle (0.2694,-1.88);
\fill[reliaink] (0.2694,-2.28) rectangle (0.3565,-1.88);
\fill[pattern=north east lines,pattern color=reachink] (0.3565,-2.28) rectangle (0.3657,-1.88);
\node[anchor=west,font=\scriptsize,inner sep=1pt] at (0.3757,-2.08) {$+0.039$ \textcolor{black!55}{16k}};
\node[anchor=east,font=\footnotesize,inner sep=1pt] at (-0.02,-2.60) {AReaL \textcolor{black!55}{\scriptsize R1-1.5B, cap 8$\to$24k, base finishes 94\%}};
\fill[finishink] (0.2222,-2.80) rectangle (0.2674,-2.40);
\fill[reliaink] (0.2674,-2.80) rectangle (0.2963,-2.40);
\fill[pattern=north east lines,pattern color=reachink] (0.2191,-2.80) rectangle (0.2222,-2.40);
\node[anchor=west,font=\scriptsize,inner sep=1pt] at (0.3063,-2.60) {$+0.019$ \textcolor{black!55}{24k}};
\node[anchor=east,font=\footnotesize,inner sep=1pt] at (-0.02,-3.12) {DeepMath \textcolor{black!55}{\scriptsize R1-1.5B, cap 24k$^\dagger$, base finishes 94\%}};
\fill[finishink] (0.2222,-3.32) rectangle (0.3139,-2.92);
\fill[reliaink] (0.3139,-3.32) rectangle (0.5077,-2.92);
\fill[pattern=north east lines,pattern color=reachink] (0.5077,-3.32) rectangle (0.5340,-2.92);
\node[anchor=west,font=\scriptsize,inner sep=1pt] at (0.5440,-3.12) {$+0.084$ \textcolor{black!55}{24k}};
\node[anchor=east,font=\footnotesize,inner sep=1pt] at (-0.02,-3.64) {DeepScaleR \textcolor{black!55}{\scriptsize R1-1.5B, cap 8$\to$24k, base finishes 94\%}};
\fill[finishink] (0.2222,-3.84) rectangle (0.2962,-3.44);
\fill[reliaink] (0.2962,-3.84) rectangle (0.3966,-3.44);
\fill[pattern=north east lines,pattern color=reachink] (0.3966,-3.84) rectangle (0.4105,-3.44);
\node[anchor=west,font=\scriptsize,inner sep=1pt] at (0.4205,-3.64) {$+0.051$ \textcolor{black!55}{24k}};
\node[anchor=east,font=\footnotesize,inner sep=1pt] at (-0.02,-4.16) {RLinf \textcolor{black!55}{\scriptsize R1-1.5B, cap 27.6k, base finishes 95\%}};
\fill[finishink] (0.2222,-4.36) rectangle (0.2854,-3.96);
\fill[reliaink] (0.2854,-4.36) rectangle (0.4198,-3.96);
\fill[pattern=north east lines,pattern color=reachink] (0.4198,-4.36) rectangle (0.4213,-3.96);
\node[anchor=west,font=\scriptsize,inner sep=1pt] at (0.4313,-4.16) {$+0.054$ \textcolor{black!55}{28k}};
\node[anchor=east,font=\footnotesize,inner sep=1pt] at (-0.02,-4.68) {STILL-3 \textcolor{black!55}{\scriptsize R1-1.5B, cap 29k$^\dagger$, base finishes 95\%}};
\fill[finishink] (0.2222,-4.88) rectangle (0.2598,-4.48);
\fill[reliaink] (0.2598,-4.88) rectangle (0.3179,-4.48);
\fill[pattern=north east lines,pattern color=reachink] (0.3179,-4.88) rectangle (0.3349,-4.48);
\node[anchor=west,font=\scriptsize,inner sep=1pt] at (0.3449,-4.68) {$+0.030$ \textcolor{black!55}{28k}};
\node[anchor=east,font=\footnotesize,inner sep=1pt] at (-0.02,-5.20) {DeepMath-Omn \textcolor{black!55}{\scriptsize OMN-1.5B, cap 24k$^\dagger$, base finishes 98\%}};
\fill[finishink] (0.2222,-5.40) rectangle (0.2312,-5.00);
\fill[reliaink] (0.2312,-5.40) rectangle (0.2994,-5.00);
\fill[pattern=north east lines,pattern color=reachink] (0.2191,-5.40) rectangle (0.2222,-5.00);
\node[anchor=west,font=\scriptsize,inner sep=1pt] at (0.3094,-5.20) {$+0.020$ \textcolor{black!55}{24k}};
\node[anchor=east,font=\footnotesize,inner sep=1pt] at (-0.02,-5.72) {KnowRL \textcolor{black!55}{\scriptsize OMN-1.5B, cap 24k, base finishes 98\%}};
\fill[finishink] (0.2222,-5.92) rectangle (0.2449,-5.52);
\fill[reliaink] (0.2449,-5.92) rectangle (0.3812,-5.52);
\node[anchor=west,font=\scriptsize,inner sep=1pt] at (0.3912,-5.72) {$+0.043$ \textcolor{black!55}{24k}};
\node[anchor=east,font=\footnotesize,inner sep=1pt] at (-0.02,-6.24) {QuestA \textcolor{black!55}{\scriptsize OMN-1.5B, cap 24k, base finishes 98\%}};
\fill[finishink] (0.0902,-6.44) rectangle (0.2222,-6.04);
\fill[reliaink] (0.2222,-6.44) rectangle (0.4113,-6.04);
\fill[pattern=north east lines,pattern color=reachink] (0.4113,-6.44) rectangle (0.4329,-6.04);
\node[anchor=west,font=\scriptsize,inner sep=1pt] at (0.4429,-6.24) {$+0.021$ \textcolor{black!55}{24k}};
\node[anchor=east,font=\footnotesize,inner sep=1pt] at (-0.02,-6.76) {Skywork-32B \textcolor{black!55}{\scriptsize R1-32B, cap 16$\to$24k, base finishes 98\%}};
\fill[finishink] (0.1903,-6.96) rectangle (0.2222,-6.56);
\fill[reliaink] (0.2222,-6.96) rectangle (0.3976,-6.56);
\fill[pattern=north east lines,pattern color=reachink] (0.3976,-6.96) rectangle (0.4470,-6.56);
\node[anchor=west,font=\scriptsize,inner sep=1pt] at (0.4570,-6.76) {$+0.052$ \textcolor{black!55}{24k}};
\node[anchor=east,font=\footnotesize,inner sep=1pt] at (-0.02,-7.28) {AceReason \textcolor{black!55}{\scriptsize R1-7B, cap 8$\to$32k, base finishes 99\%}};
\fill[finishink] (0.2222,-7.48) rectangle (0.2516,-7.08);
\fill[reliaink] (0.2516,-7.48) rectangle (0.3318,-7.08);
\fill[pattern=north east lines,pattern color=reachink] (0.2145,-7.48) rectangle (0.2222,-7.08);
\node[anchor=west,font=\scriptsize,inner sep=1pt] at (0.3418,-7.28) {$+0.028$ \textcolor{black!55}{32k}};
\node[anchor=east,font=\footnotesize,inner sep=1pt] at (-0.02,-7.80) {Skywork-7B \textcolor{black!55}{\scriptsize R1-7B, cap 16$\to$32k, base finishes 99\%}};
\fill[finishink] (0.1925,-8.00) rectangle (0.2222,-7.60);
\fill[reliaink] (0.2222,-8.00) rectangle (0.4078,-7.60);
\fill[pattern=north east lines,pattern color=reachink] (0.4078,-8.00) rectangle (0.4417,-7.60);
\node[anchor=west,font=\scriptsize,inner sep=1pt] at (0.4517,-7.80) {$+0.051$ \textcolor{black!55}{32k}};
\node[anchor=east,font=\footnotesize,inner sep=1pt] at (-0.02,-8.32) {Skywork-Math \textcolor{black!55}{\scriptsize R1-7B, cap 8$\to$32k, base finishes 99\%}};
\fill[finishink] (0.2222,-8.52) rectangle (0.2262,-8.12);
\fill[reliaink] (0.2262,-8.52) rectangle (0.3580,-8.12);
\fill[pattern=north east lines,pattern color=reachink] (0.3580,-8.52) rectangle (0.4059,-8.12);
\node[anchor=west,font=\scriptsize,inner sep=1pt] at (0.4159,-8.32) {$+0.050$ \textcolor{black!55}{32k}};
\node[anchor=east,font=\footnotesize,inner sep=1pt] at (-0.02,-8.84) {AceReason-14B \textcolor{black!55}{\scriptsize R1-14B, cap 8$\to$32k, base finishes 99\%}};
\fill[finishink] (0.2222,-9.04) rectangle (0.2443,-8.64);
\fill[reliaink] (0.2443,-9.04) rectangle (0.4105,-8.64);
\fill[pattern=north east lines,pattern color=reachink] (0.4105,-9.04) rectangle (0.4491,-8.64);
\node[anchor=west,font=\scriptsize,inner sep=1pt] at (0.4591,-8.84) {$+0.061$ \textcolor{black!55}{32k}};
\node[anchor=east,font=\footnotesize,inner sep=1pt] at (-0.02,-9.36) {Polaris-1.7B \textcolor{black!55}{\scriptsize Q3-1.7B, cap 40$\to$48k, base finishes 99\%}};
\fill[finishink] (0.1229,-9.56) rectangle (0.2222,-9.16);
\fill[reliaink] (0.2222,-9.56) rectangle (0.3076,-9.16);
\fill[pattern=north east lines,pattern color=reachink] (0.3076,-9.56) rectangle (0.3200,-9.16);
\node[anchor=west,font=\scriptsize,inner sep=1pt] at (0.3300,-9.36) {$0$ \textcolor{black!55}{48k}};
\node[anchor=east,font=\footnotesize,inner sep=1pt] at (-0.02,-9.88) {Polaris-4B \textcolor{black!55}{\scriptsize Q3-4B, cap 40$\to$52k, base finishes 100\%}};
\fill[finishink] (0.1613,-10.08) rectangle (0.2222,-9.68);
\fill[reliaink] (0.2222,-10.08) rectangle (0.3001,-9.68);
\fill[pattern=north east lines,pattern color=reachink] (0.3001,-10.08) rectangle (0.3170,-9.68);
\node[anchor=west,font=\scriptsize,inner sep=1pt] at (0.3270,-9.88) {$+0.009$ \textcolor{black!55}{52k}};
\draw[black!60,thin] (0.2222,0.21) -- (0.2222,-10.28);
\draw[black!50,thin] (0,-10.28) -- (1,-10.28);
\foreach \v/\l in {-0.05/$-0.05$,0/$0$,0.05/$0.05$,0.1/$0.10$,0.15/$0.15$,0.2/$0.20$} \node[anchor=north,font=\scriptsize] at ({(\v+0.06)*3.7037},-10.30) {\l};
\node[anchor=north,font=\scriptsize] at (0.5,-10.58) {change in pass@1 over the base, physics};
\end{tikzpicture}\hspace{1.1cm}\begin{tikzpicture}[x=0.8cm,y=0.66cm,baseline=0pt]
\node[font=\small] at (0,0.47) {at the cap, wins alone};
\fill[black!35] (0.000,-0.20) rectangle (0,0.20);
\fill[black] (0,-0.20) rectangle (0.000,0.20);
\node[anchor=east,font=\scriptsize,inner sep=1pt] at (-0.030,0.00) {0};
\node[anchor=west,font=\scriptsize,inner sep=1pt] at (0.030,0.00) {0};
\fill[black!35] (-0.455,-0.72) rectangle (0,-0.32);
\fill[black] (0,-0.72) rectangle (0.136,-0.32);
\node[anchor=east,font=\scriptsize,inner sep=1pt] at (-0.485,-0.52) {10};
\node[anchor=west,font=\scriptsize,inner sep=1pt] at (0.166,-0.52) {3};
\fill[black!35] (-0.636,-1.24) rectangle (0,-0.84);
\fill[black] (0,-1.24) rectangle (0.364,-0.84);
\node[anchor=east,font=\scriptsize,inner sep=1pt] at (-0.666,-1.04) {14};
\node[anchor=west,font=\scriptsize,inner sep=1pt] at (0.394,-1.04) {8};
\fill[black!35] (-0.591,-1.76) rectangle (0,-1.36);
\fill[black] (0,-1.76) rectangle (0.318,-1.36);
\node[anchor=east,font=\scriptsize,inner sep=1pt] at (-0.621,-1.56) {13};
\node[anchor=west,font=\scriptsize,inner sep=1pt] at (0.348,-1.56) {7};
\fill[black!35] (-0.500,-2.28) rectangle (0,-1.88);
\fill[black] (0,-2.28) rectangle (0.682,-1.88);
\node[anchor=east,font=\scriptsize,inner sep=1pt] at (-0.530,-2.08) {11};
\node[anchor=west,font=\scriptsize,inner sep=1pt] at (0.712,-2.08) {15};
\fill[black!35] (-0.636,-2.80) rectangle (0,-2.40);
\fill[black] (0,-2.80) rectangle (0.545,-2.40);
\node[anchor=east,font=\scriptsize,inner sep=1pt] at (-0.666,-2.60) {14};
\node[anchor=west,font=\scriptsize,inner sep=1pt] at (0.575,-2.60) {12};
\fill[black!35] (-0.455,-3.32) rectangle (0,-2.92);
\fill[black] (0,-3.32) rectangle (0.591,-2.92);
\node[anchor=east,font=\scriptsize,inner sep=1pt] at (-0.485,-3.12) {10};
\node[anchor=west,font=\scriptsize,inner sep=1pt] at (0.621,-3.12) {13};
\fill[black!35] (-0.773,-3.84) rectangle (0,-3.44);
\fill[black] (0,-3.84) rectangle (0.773,-3.44);
\node[anchor=east,font=\scriptsize,inner sep=1pt] at (-0.803,-3.64) {17};
\node[anchor=west,font=\scriptsize,inner sep=1pt] at (0.803,-3.64) {17};
\fill[black!35] (-1.000,-4.36) rectangle (0,-3.96);
\fill[black] (0,-4.36) rectangle (0.818,-3.96);
\node[anchor=east,font=\scriptsize,inner sep=1pt] at (-1.030,-4.16) {22};
\node[anchor=west,font=\scriptsize,inner sep=1pt] at (0.848,-4.16) {18};
\fill[black!35] (-0.409,-4.88) rectangle (0,-4.48);
\fill[black] (0,-4.88) rectangle (0.636,-4.48);
\node[anchor=east,font=\scriptsize,inner sep=1pt] at (-0.439,-4.68) {9};
\node[anchor=west,font=\scriptsize,inner sep=1pt] at (0.666,-4.68) {14};
\fill[black!35] (-0.591,-5.40) rectangle (0,-5.00);
\fill[black] (0,-5.40) rectangle (0.545,-5.00);
\node[anchor=east,font=\scriptsize,inner sep=1pt] at (-0.621,-5.20) {13};
\node[anchor=west,font=\scriptsize,inner sep=1pt] at (0.575,-5.20) {12};
\fill[black!35] (-0.864,-5.92) rectangle (0,-5.52);
\fill[black] (0,-5.92) rectangle (0.955,-5.52);
\node[anchor=east,font=\scriptsize,inner sep=1pt] at (-0.894,-5.72) {19};
\node[anchor=west,font=\scriptsize,inner sep=1pt] at (0.985,-5.72) {21};
\fill[black!35] (-0.227,-6.44) rectangle (0,-6.04);
\fill[black] (0,-6.44) rectangle (0.773,-6.04);
\node[anchor=east,font=\scriptsize,inner sep=1pt] at (-0.257,-6.24) {5};
\node[anchor=west,font=\scriptsize,inner sep=1pt] at (0.803,-6.24) {17};
\fill[black!35] (-0.636,-6.96) rectangle (0,-6.56);
\fill[black] (0,-6.96) rectangle (0.727,-6.56);
\node[anchor=east,font=\scriptsize,inner sep=1pt] at (-0.666,-6.76) {14};
\node[anchor=west,font=\scriptsize,inner sep=1pt] at (0.757,-6.76) {16};
\fill[black!35] (-0.864,-7.48) rectangle (0,-7.08);
\fill[black] (0,-7.48) rectangle (0.773,-7.08);
\node[anchor=east,font=\scriptsize,inner sep=1pt] at (-0.894,-7.28) {19};
\node[anchor=west,font=\scriptsize,inner sep=1pt] at (0.803,-7.28) {17};
\fill[black!35] (-0.682,-8.00) rectangle (0,-7.60);
\fill[black] (0,-8.00) rectangle (0.864,-7.60);
\node[anchor=east,font=\scriptsize,inner sep=1pt] at (-0.712,-7.80) {15};
\node[anchor=west,font=\scriptsize,inner sep=1pt] at (0.894,-7.80) {19};
\fill[black!35] (-0.727,-8.52) rectangle (0,-8.12);
\fill[black] (0,-8.52) rectangle (0.818,-8.12);
\node[anchor=east,font=\scriptsize,inner sep=1pt] at (-0.757,-8.32) {16};
\node[anchor=west,font=\scriptsize,inner sep=1pt] at (0.848,-8.32) {18};
\fill[black!35] (-0.545,-9.04) rectangle (0,-8.64);
\fill[black] (0,-9.04) rectangle (0.955,-8.64);
\node[anchor=east,font=\scriptsize,inner sep=1pt] at (-0.575,-8.84) {12};
\node[anchor=west,font=\scriptsize,inner sep=1pt] at (0.985,-8.84) {21};
\fill[black!35] (-0.500,-9.56) rectangle (0,-9.16);
\fill[black] (0,-9.56) rectangle (0.773,-9.16);
\node[anchor=east,font=\scriptsize,inner sep=1pt] at (-0.530,-9.36) {11};
\node[anchor=west,font=\scriptsize,inner sep=1pt] at (0.803,-9.36) {17};
\fill[black!35] (-0.682,-10.08) rectangle (0,-9.68);
\fill[black] (0,-10.08) rectangle (0.636,-9.68);
\node[anchor=east,font=\scriptsize,inner sep=1pt] at (-0.712,-9.88) {15};
\node[anchor=west,font=\scriptsize,inner sep=1pt] at (0.666,-9.88) {14};
\draw[black!60,thin] (0,0.21) -- (0,-10.28);
\node[anchor=north,font=\scriptsize] at (-0.75,-10.30) {base only};
\node[anchor=north,font=\scriptsize] at (0.75,-10.30) {RL only};
\end{tikzpicture}\hspace{0.5cm}\begin{tikzpicture}[x=4.20cm,y=0.66cm,baseline=0pt]
\node[font=\small] at (0.5,0.47) {next budget above the cap};
\fill[finishink] (0.2222,-0.20) rectangle (0.2290,0.20);
\fill[reliaink] (0.2047,-0.20) rectangle (0.2222,0.20);
\fill[pattern=north east lines,pattern color=reachink] (0.2016,-0.20) rectangle (0.2047,0.20);
\node[anchor=west,font=\scriptsize,inner sep=1pt] at (0.2390,0.00) {$-0.004$ \textcolor{black!55}{8k}};
\fill[finishink] (0.2222,-0.72) rectangle (0.4997,-0.32);
\fill[reliaink] (0.4997,-0.72) rectangle (0.6960,-0.32);
\fill[pattern=north east lines,pattern color=reachink] (0.6960,-0.72) rectangle (0.7068,-0.32);
\node[anchor=west,font=\scriptsize,inner sep=1pt] at (0.7168,-0.52) {$+0.131$ \textcolor{black!55}{16k}};
\fill[finishink] (0.2222,-1.24) rectangle (0.3205,-0.84);
\fill[reliaink] (0.3205,-1.24) rectangle (0.5278,-0.84);
\fill[pattern=north east lines,pattern color=reachink] (0.5278,-1.24) rectangle (0.5432,-0.84);
\node[anchor=west,font=\scriptsize,inner sep=1pt] at (0.5532,-1.04) {$+0.087$ \textcolor{black!55}{24k}};
\fill[finishink] (0.2222,-1.76) rectangle (0.2647,-1.36);
\fill[reliaink] (0.2647,-1.76) rectangle (0.2793,-1.36);
\fill[pattern=north east lines,pattern color=reachink] (0.2793,-1.76) rectangle (0.2855,-1.36);
\node[anchor=west,font=\scriptsize,inner sep=1pt] at (0.2955,-1.56) {$+0.017$ \textcolor{black!55}{24k}};
\fill[finishink] (0.2222,-2.28) rectangle (0.2546,-1.88);
\fill[reliaink] (0.2546,-2.28) rectangle (0.3441,-1.88);
\fill[pattern=north east lines,pattern color=reachink] (0.3441,-2.28) rectangle (0.3781,-1.88);
\node[anchor=west,font=\scriptsize,inner sep=1pt] at (0.3881,-2.08) {$+0.042$ \textcolor{black!55}{24k}};
\fill[finishink] (0.2222,-2.80) rectangle (0.2674,-2.40);
\fill[reliaink] (0.2674,-2.80) rectangle (0.3025,-2.40);
\fill[pattern=north east lines,pattern color=reachink] (0.2099,-2.80) rectangle (0.2222,-2.40);
\node[anchor=west,font=\scriptsize,inner sep=1pt] at (0.3125,-2.60) {$+0.018$ \textcolor{black!55}{28k}};
\fill[finishink] (0.2222,-3.32) rectangle (0.2997,-2.92);
\fill[reliaink] (0.2997,-3.32) rectangle (0.4954,-2.92);
\fill[pattern=north east lines,pattern color=reachink] (0.4954,-3.32) rectangle (0.5309,-2.92);
\node[anchor=west,font=\scriptsize,inner sep=1pt] at (0.5409,-3.12) {$+0.083$ \textcolor{black!55}{28k}};
\fill[finishink] (0.2222,-3.84) rectangle (0.2814,-3.44);
\fill[reliaink] (0.2814,-3.84) rectangle (0.3873,-3.44);
\fill[pattern=north east lines,pattern color=reachink] (0.3873,-3.84) rectangle (0.4059,-3.44);
\node[anchor=west,font=\scriptsize,inner sep=1pt] at (0.4159,-3.64) {$+0.050$ \textcolor{black!55}{28k}};
\fill[finishink] (0.2222,-4.36) rectangle (0.2730,-3.96);
\fill[reliaink] (0.2730,-4.36) rectangle (0.4074,-3.96);
\fill[pattern=north east lines,pattern color=reachink] (0.4074,-4.36) rectangle (0.4198,-3.96);
\node[anchor=west,font=\scriptsize,inner sep=1pt] at (0.4298,-4.16) {$+0.053$ \textcolor{black!55}{32k}};
\fill[finishink] (0.2222,-4.88) rectangle (0.2515,-4.48);
\fill[reliaink] (0.2515,-4.88) rectangle (0.3086,-4.48);
\fill[pattern=north east lines,pattern color=reachink] (0.3086,-4.88) rectangle (0.3333,-4.48);
\node[anchor=west,font=\scriptsize,inner sep=1pt] at (0.3433,-4.68) {$+0.030$ \textcolor{black!55}{32k}};
\fill[finishink] (0.2222,-5.40) rectangle (0.2353,-5.00);
\fill[reliaink] (0.2353,-5.40) rectangle (0.3009,-5.00);
\fill[pattern=north east lines,pattern color=reachink] (0.2191,-5.40) rectangle (0.2222,-5.00);
\node[anchor=west,font=\scriptsize,inner sep=1pt] at (0.3109,-5.20) {$+0.020$ \textcolor{black!55}{28k}};
\fill[finishink] (0.2222,-5.92) rectangle (0.2540,-5.52);
\fill[reliaink] (0.2540,-5.92) rectangle (0.3827,-5.52);
\fill[pattern=north east lines,pattern color=reachink] (0.3827,-5.92) rectangle (0.3843,-5.52);
\node[anchor=west,font=\scriptsize,inner sep=1pt] at (0.3943,-5.72) {$+0.044$ \textcolor{black!55}{28k}};
\fill[finishink] (0.1595,-6.44) rectangle (0.2222,-6.04);
\fill[reliaink] (0.2222,-6.44) rectangle (0.3852,-6.04);
\fill[pattern=north east lines,pattern color=reachink] (0.3852,-6.44) rectangle (0.3945,-6.04);
\node[anchor=west,font=\scriptsize,inner sep=1pt] at (0.4045,-6.24) {$+0.030$ \textcolor{black!55}{28k}};
\fill[finishink] (0.2188,-6.96) rectangle (0.2222,-6.56);
\fill[reliaink] (0.2222,-6.96) rectangle (0.3954,-6.56);
\fill[pattern=north east lines,pattern color=reachink] (0.3954,-6.96) rectangle (0.4432,-6.56);
\node[anchor=west,font=\scriptsize,inner sep=1pt] at (0.4532,-6.76) {$+0.059$ \textcolor{black!55}{28k}};
\fill[finishink] (0.2222,-7.48) rectangle (0.2310,-7.08);
\fill[reliaink] (0.2310,-7.48) rectangle (0.3210,-7.08);
\fill[pattern=north east lines,pattern color=reachink] (0.2145,-7.48) rectangle (0.2222,-7.08);
\node[anchor=west,font=\scriptsize,inner sep=1pt] at (0.3310,-7.28) {$+0.025$ \textcolor{black!55}{48k}};
\fill[finishink] (0.2202,-8.00) rectangle (0.2222,-7.60);
\fill[reliaink] (0.2222,-8.00) rectangle (0.3863,-7.60);
\fill[pattern=north east lines,pattern color=reachink] (0.3863,-8.00) rectangle (0.4249,-7.60);
\node[anchor=west,font=\scriptsize,inner sep=1pt] at (0.4349,-7.80) {$+0.054$ \textcolor{black!55}{48k}};
\fill[finishink] (0.2222,-8.52) rectangle (0.2282,-8.12);
\fill[reliaink] (0.2282,-8.52) rectangle (0.3426,-8.12);
\fill[pattern=north east lines,pattern color=reachink] (0.3426,-8.52) rectangle (0.4012,-8.12);
\node[anchor=west,font=\scriptsize,inner sep=1pt] at (0.4112,-8.32) {$+0.048$ \textcolor{black!55}{48k}};
\fill[finishink] (0.2222,-9.04) rectangle (0.2466,-8.64);
\fill[reliaink] (0.2466,-9.04) rectangle (0.4105,-8.64);
\fill[pattern=north east lines,pattern color=reachink] (0.4105,-9.04) rectangle (0.4506,-8.64);
\node[anchor=west,font=\scriptsize,inner sep=1pt] at (0.4606,-8.84) {$+0.062$ \textcolor{black!55}{48k}};
\fill[finishink] (0.1359,-9.56) rectangle (0.2222,-9.16);
\fill[reliaink] (0.2222,-9.56) rectangle (0.3070,-9.16);
\fill[pattern=north east lines,pattern color=reachink] (0.3070,-9.56) rectangle (0.3086,-9.16);
\node[anchor=west,font=\scriptsize,inner sep=1pt] at (0.3186,-9.36) {$0$ \textcolor{black!55}{52k}};
\fill[finishink] (0.1672,-10.08) rectangle (0.2222,-9.68);
\fill[reliaink] (0.2222,-10.08) rectangle (0.2988,-9.68);
\fill[pattern=north east lines,pattern color=reachink] (0.2988,-10.08) rectangle (0.3158,-9.68);
\node[anchor=west,font=\scriptsize,inner sep=1pt] at (0.3258,-9.88) {$+0.010$ \textcolor{black!55}{64k}};
\draw[black!60,thin] (0.2222,0.21) -- (0.2222,-10.28);
\draw[black!50,thin] (0,-10.28) -- (1,-10.28);
\foreach \v/\l in {-0.05/$-0.05$,0/$0$,0.05/$0.05$,0.1/$0.10$,0.15/$0.15$,0.2/$0.20$} \node[anchor=north,font=\scriptsize] at ({(\v+0.06)*3.7037},-10.30) {\l};
\node[anchor=north,font=\scriptsize] at (0.5,-10.58) {change in pass@1 over the base, physics};
\end{tikzpicture}\hspace{1.1cm}\begin{tikzpicture}[x=0.8cm,y=0.66cm,baseline=0pt]
\node[font=\small] at (0,0.47) {above the cap, wins alone};
\fill[black!35] (-0.125,-0.20) rectangle (0,0.20);
\fill[black] (0,-0.20) rectangle (0.000,0.20);
\node[anchor=east,font=\scriptsize,inner sep=1pt] at (-0.155,0.00) {3};
\node[anchor=west,font=\scriptsize,inner sep=1pt] at (0.030,0.00) {0};
\fill[black!35] (-0.708,-0.72) rectangle (0,-0.32);
\fill[black] (0,-0.72) rectangle (0.375,-0.32);
\node[anchor=east,font=\scriptsize,inner sep=1pt] at (-0.738,-0.52) {17};
\node[anchor=west,font=\scriptsize,inner sep=1pt] at (0.405,-0.52) {9};
\fill[black!35] (-0.667,-1.24) rectangle (0,-0.84);
\fill[black] (0,-1.24) rectangle (0.750,-0.84);
\node[anchor=east,font=\scriptsize,inner sep=1pt] at (-0.697,-1.04) {16};
\node[anchor=west,font=\scriptsize,inner sep=1pt] at (0.780,-1.04) {18};
\fill[black!35] (-0.750,-1.76) rectangle (0,-1.36);
\fill[black] (0,-1.76) rectangle (0.708,-1.36);
\node[anchor=east,font=\scriptsize,inner sep=1pt] at (-0.780,-1.56) {18};
\node[anchor=west,font=\scriptsize,inner sep=1pt] at (0.738,-1.56) {17};
\fill[black!35] (-0.583,-2.28) rectangle (0,-1.88);
\fill[black] (0,-2.28) rectangle (0.833,-1.88);
\node[anchor=east,font=\scriptsize,inner sep=1pt] at (-0.613,-2.08) {14};
\node[anchor=west,font=\scriptsize,inner sep=1pt] at (0.863,-2.08) {20};
\fill[black!35] (-0.792,-2.80) rectangle (0,-2.40);
\fill[black] (0,-2.80) rectangle (0.583,-2.40);
\node[anchor=east,font=\scriptsize,inner sep=1pt] at (-0.822,-2.60) {19};
\node[anchor=west,font=\scriptsize,inner sep=1pt] at (0.613,-2.60) {14};
\fill[black!35] (-0.417,-3.32) rectangle (0,-2.92);
\fill[black] (0,-3.32) rectangle (0.708,-2.92);
\node[anchor=east,font=\scriptsize,inner sep=1pt] at (-0.447,-3.12) {10};
\node[anchor=west,font=\scriptsize,inner sep=1pt] at (0.738,-3.12) {17};
\fill[black!35] (-0.708,-3.84) rectangle (0,-3.44);
\fill[black] (0,-3.84) rectangle (0.833,-3.44);
\node[anchor=east,font=\scriptsize,inner sep=1pt] at (-0.738,-3.64) {17};
\node[anchor=west,font=\scriptsize,inner sep=1pt] at (0.863,-3.64) {20};
\fill[black!35] (-1.000,-4.36) rectangle (0,-3.96);
\fill[black] (0,-4.36) rectangle (0.917,-3.96);
\node[anchor=east,font=\scriptsize,inner sep=1pt] at (-1.030,-4.16) {24};
\node[anchor=west,font=\scriptsize,inner sep=1pt] at (0.947,-4.16) {22};
\fill[black!35] (-0.375,-4.88) rectangle (0,-4.48);
\fill[black] (0,-4.88) rectangle (0.750,-4.48);
\node[anchor=east,font=\scriptsize,inner sep=1pt] at (-0.405,-4.68) {9};
\node[anchor=west,font=\scriptsize,inner sep=1pt] at (0.780,-4.68) {18};
\fill[black!35] (-0.542,-5.40) rectangle (0,-5.00);
\fill[black] (0,-5.40) rectangle (0.500,-5.00);
\node[anchor=east,font=\scriptsize,inner sep=1pt] at (-0.572,-5.20) {13};
\node[anchor=west,font=\scriptsize,inner sep=1pt] at (0.530,-5.20) {12};
\fill[black!35] (-0.792,-5.92) rectangle (0,-5.52);
\fill[black] (0,-5.92) rectangle (0.917,-5.52);
\node[anchor=east,font=\scriptsize,inner sep=1pt] at (-0.822,-5.72) {19};
\node[anchor=west,font=\scriptsize,inner sep=1pt] at (0.947,-5.72) {22};
\fill[black!35] (-0.417,-6.44) rectangle (0,-6.04);
\fill[black] (0,-6.44) rectangle (0.750,-6.04);
\node[anchor=east,font=\scriptsize,inner sep=1pt] at (-0.447,-6.24) {10};
\node[anchor=west,font=\scriptsize,inner sep=1pt] at (0.780,-6.24) {18};
\fill[black!35] (-0.625,-6.96) rectangle (0,-6.56);
\fill[black] (0,-6.96) rectangle (0.708,-6.56);
\node[anchor=east,font=\scriptsize,inner sep=1pt] at (-0.655,-6.76) {15};
\node[anchor=west,font=\scriptsize,inner sep=1pt] at (0.738,-6.76) {17};
\fill[black!35] (-0.792,-7.48) rectangle (0,-7.08);
\fill[black] (0,-7.48) rectangle (0.708,-7.08);
\node[anchor=east,font=\scriptsize,inner sep=1pt] at (-0.822,-7.28) {19};
\node[anchor=west,font=\scriptsize,inner sep=1pt] at (0.738,-7.28) {17};
\fill[black!35] (-0.667,-8.00) rectangle (0,-7.60);
\fill[black] (0,-8.00) rectangle (0.917,-7.60);
\node[anchor=east,font=\scriptsize,inner sep=1pt] at (-0.697,-7.80) {16};
\node[anchor=west,font=\scriptsize,inner sep=1pt] at (0.947,-7.80) {22};
\fill[black!35] (-0.708,-8.52) rectangle (0,-8.12);
\fill[black] (0,-8.52) rectangle (0.833,-8.12);
\node[anchor=east,font=\scriptsize,inner sep=1pt] at (-0.738,-8.32) {17};
\node[anchor=west,font=\scriptsize,inner sep=1pt] at (0.863,-8.32) {20};
\fill[black!35] (-0.500,-9.04) rectangle (0,-8.64);
\fill[black] (0,-9.04) rectangle (0.917,-8.64);
\node[anchor=east,font=\scriptsize,inner sep=1pt] at (-0.530,-8.84) {12};
\node[anchor=west,font=\scriptsize,inner sep=1pt] at (0.947,-8.84) {22};
\fill[black!35] (-0.500,-9.56) rectangle (0,-9.16);
\fill[black] (0,-9.56) rectangle (0.708,-9.16);
\node[anchor=east,font=\scriptsize,inner sep=1pt] at (-0.530,-9.36) {12};
\node[anchor=west,font=\scriptsize,inner sep=1pt] at (0.738,-9.36) {17};
\fill[black!35] (-0.625,-10.08) rectangle (0,-9.68);
\fill[black] (0,-10.08) rectangle (0.583,-9.68);
\node[anchor=east,font=\scriptsize,inner sep=1pt] at (-0.655,-9.88) {15};
\node[anchor=west,font=\scriptsize,inner sep=1pt] at (0.613,-9.88) {14};
\draw[black!60,thin] (0,0.21) -- (0,-10.28);
\node[anchor=north,font=\scriptsize] at (-0.75,-10.30) {base only};
\node[anchor=north,font=\scriptsize] at (0.75,-10.30) {RL only};
\end{tikzpicture}}\\[3pt]
{\scriptsize \tikz\fill[finishink] (0,0) rectangle (0.3,0.18); finishing term $F$\quad \tikz\fill[reliaink] (0,0) rectangle (0.3,0.18); reliability term $P$\quad \tikz\fill[pattern=north east lines,pattern color=reachink] (0,0) rectangle (0.3,0.18); reach, net of its mirror}
\end{minipage}
\caption{What outcome RL buys at each recipe's own training cap and at the next budget above it, in pass@1 on physics. Each row is one recipe, split as in Figure~\ref{fig:overview}. Parts left of zero read as in Figure~\ref{fig:overview}. The number at the right of each row is the change, with the budget read. Red \emph{parts} $<0.005$ marks a row whose parts are all below $0.005$, too small to read. Rows are sorted by how often the base finishes at the cap on physics, least first. The lone-win panels count physics problems that only the RL model solves (black, right) or only the base (grey, left), while the other model finished all eight samples. $^\dagger$: the cap is not exact. DeepMath trains at a $24$k cap but penalizes length from $20$k. STILL-3 trains at $29$k and is read at $28$k.}
\label{fig:recipes}
\end{figure*}
Outcome RL, which rewards only a correct final answer, is a standard way to train reasoning models. Its gains are reported as pass@1. Whether such a gain is new knowledge or changed behaviour matters. New knowledge widens what a model can solve. Changed behaviour only helps it use what it already knows. The two are hard to tell apart. Every pass@1 is scored under a token budget, so one number mixes answering in time, answering right and solving new problems. \citet{yue2025} found that outcome RL adds no reasoning the base cannot already produce. Pass@$k$ is the chance that at least one of $k$ samples is right. Once $k$ reaches the hundreds, the base scores higher than the RL model. The problems the RL model solves are nearly a subset of those its base solves. Prior work calls this narrowing \emph{sharpening} \citep{huang2024sharpening,qin2025decomposing,wu2025knowledge}. We ask whether RL changes behaviour or knowledge at $k=1$, the score every report rests on. Finishing in time and being right more often change behaviour. Solving problems the base cannot points to new knowledge. We count a gain as new knowledge only when the base finishes all eight samples on the problem and gets every one wrong. We count any other gain as changed behaviour, since the base either answers the problem right at least once or does not finish every sample.

Our first question asks of released models: how much of a pass@1 gain is finishing within the token budget, and is the rest reliability or reach? Every recipe is also trained under a budget. If part of a gain is the model fitting its answer into the budget, comparisons between recipes, or between a recipe and its base, partly measure how long each model writes rather than how well it reasons. We ask the first question of thinking models on physics and math. A bounded second question follows: can training install knowledge the model does not use on its own, when it is supplied?

Our main study answers the first question by reading twenty released RL models against their bases at fixed budgets. All twenty start from thinking bases, which distillation has already widened. \citet{yue2025} leave thinking bases open.   Once the budget suffices, reliability rises on physics wherever the base already finishes. Reach stays small.

The second question tests the obvious remedy for sharpening: hand the model the knowledge its derivations lack (a hint, a teacher's derivation, or a question about the setup). We train an $8$B thinking model with outcome RL on physics, supplying in five ways the physical setup of problems its base almost never solves. At this small scale, the supplied setup changes the model during training but is not shown to carry over to held-out problems beyond plain RL. Reward raises a problem's solve rate only once the model samples a success on that problem (\S\ref{sec:install}). Released recipes that use the remedy, QuestA and KnowRL, reach only a little further. Their net reach is at most $0.018$ (\S\ref{sec:reach}).

\paragraph{Contributions.} (1) \emph{A budget-aware split of the pass@1 gain} into finishing, reliability and reach (Eqs.~\eqref{eq:finrel}--\eqref{eq:robustreach}). Unlike pass@$k$, it separates answering in time from answering right. The split also nets reach against the reverse. (2) \emph{A read of twenty released pairs on thinking bases}, the case \citet{yue2025} leave open, at each recipe's own training cap and the next budget above it (\S\ref{sec:what}). (3) \emph{A comparison rule}: compare a recipe with its base at or above its training cap. (4) \emph{A controlled probe} that supplies an $8$B model, in five ways, the setup of physics problems its base almost never solves, and tests transfer to their text-only twins (\S\ref{sec:install}).

\section{Related Work}
\label{sec:related}

\textbf{What RL with verifiable rewards changes.} Prior work disagrees on what outcome RL changes. None splits the measured gain into finishing and the rest. On pass@$k$ curves, RL raises pass@1, but the base wins at large $k$ \citep{yue2025,wu2025leash}. RL stays within what the base can sample. \citet{yue2025} count the problems only the RL model solves at $0$--$1\%$ on AIME24 and MATH500, at $k$ up to $1{,}024$. Follow-ups trace this lost coverage to hard prompts whose rare successes drop out of small sample groups \citep{zhou2026boundary} and to winner-take-all reinforcement of what the base already finds likely \citep{nguyen2025paradox}. Others trace part of reported gains to budget mismatch and attempt inflation \citep{wu2025hidden,yang2026kernel}. Studies of derivations say RL narrows the policy where fine-tuning broadens it \citep{squeezes2025}, acts on a few tokens where derivations branch \citep{wang2025forking}, and amplifies habits the base already has \citep{gandhi2025cognitive}. Power sampling, entropy minimisation and even spurious rewards reproduce much of RL's gain \citep{powersampling2025,agarwal2025em,shao2025spurious}. One training example recovers most of a full data set's gain \citep{wang2025oneshot}. A second RL stage shortens derivations at no cost in accuracy \citep{fatemi2025concise}. Both results say RL moves behaviour more than content. Prolonged training, by contrast, is reported to expand the solvable set \citep{prorl2025}. Decompositions find RL improves execution within a fixed coverage \citep{qin2025decomposing}. Fine-tuning, not RL, moves knowledge \citep{wu2025knowledge}. All this evidence is observational: no work supplies the missing content and tests whether RL learns it. Our reliability term is the execution gain of \citet{qin2025decomposing} with finishing removed. We compare reliability with finishing at each budget. Our reach is a term of an exact split of the pass@1 gain, counted per problem at $k=8$. The base-only count in reach controls for sampling noise. Problems only one model solves are resampled and read. Unlike budget corrections of reported gains \citep{wu2025hidden}, we split the gain at a fixed budget. We also read each recipe at its own cap.

\textbf{Objectives, samplers and replay.} Other work changes how RL credits rare successes. We use one such method, replay. Pass@$k$ objectives reshape the group advantage \citep{passk2025,pkpo2025,drgrpo}. VinePPO branches continuations to estimate credit \citep{vineppo}. RLEP replays past successes into the group \citep{rlep2025}. Replay shows what reward teaches once a success is sampled.

\textbf{Measuring RL on thinking models.} Length and answer rate are known confounds \citep{drgrpo,dapo,l1_2025,thinkprune2025}. For a fixed thinking model, accuracy at a cap rises with the cap \citep{s1,qwen3}. Any change to derivation length therefore changes the score. We separate the two with the finishing decomposition. We also fix a held-out set before training and measure its noise floor.

\section{Experimental Setup}
\label{sec:setup}

This section gives the study's scope, the three metrics that split pass@1 under a budget (\S\ref{sec:metrics}) and the protocol for reading the twenty released pairs (\S\ref{sec:released}).

\textbf{Scope: thinking bases.} We study thinking bases because they give RL the most to change: how long the model reasons, when it commits to an answer and how often that answer is right, not only whether it answers. A token budget makes each change visible in the score. Practice uses thinking bases too. The strongest open models on math and science, DeepSeek-R1 and its distills and Qwen3, ship in this thinking mode. Every recipe we study starts from one of them. All models studied are thinking models: the five bases (DeepSeek-R1-Distill-Qwen 1.5B and 7B, Qwen3-1.7B and 4B in thinking mode, OpenMath-Nemotron-1.5B), the $14$B and $32$B bases added later, and our controlled run's Qwen3-VL-8B-Thinking. Models that answer in a few hundred tokens are out of scope, except one pair read for contrast: Qwen2.5-7B and its outcome-RL version, Open-Reasoner-Zero-7B (\S\ref{sec:termination}).

\textbf{Scope: physics and math.} To tell changed behaviour from new knowledge, we read every pair in two domains. Math is the recipes' own domain: all twenty recipes we study train on competition math and report math results. A gain on math shows what RL buys where it was trained. A gain on physics shows what RL buys outside that domain. Physics is at most about $1\%$ of the RL training data of every recipe that states its mix, except ProRL v2, whose mix includes STEM. Skywork-OR1-Math-7B does not state its mix. Physics problems also need modelling, which math problems rarely exercise. A physics derivation first picks the governing principle and decides which quantities and effects matter, and only then does the algebra. A wrong setup is not repaired by careful algebra later. Once they finish at $16$k, the five bases are right on $85$--$91\%$ of their MATH-500 derivations but only $21$--$56\%$ of their physics ones. The modelling also makes physics derivations long, so the token budget binds more often than on math (\S\ref{sec:termination}).

\textbf{Test sets.} Each domain has a main set and a check. The main physics set is $300$ text-only problems from the training corpus of our controlled run (\S\ref{sec:install}), so both questions share one audited answer checker. The set keeps only problems the controlled run's base solves at least sometimes, so pass@1 can move. Filtering on the base caps how much reach can show, so $671$ unfiltered physics problems check the reach result (Appendix~\ref{sec:hardset}). On math, the main set is MATH-500, the recipes' standard benchmark. The math check is $276$ recent contest problems, below MATH-500's ceiling, some released after every model (Appendix~\ref{sec:mathset}). We call it competition math.

\begin{figure}[!t]
\centering
\definecolor{finishink}{RGB}{16,50,88}
\definecolor{reliaink}{RGB}{242,146,24}
\definecolor{reachink}{RGB}{0,158,115}
\resizebox{\linewidth}{!}{%
\begin{tikzpicture}[x=4.60cm,y=0.80cm]
\node[anchor=east,font=\scriptsize,inner sep=1pt] at (-0.02,0.00) {Open-RS3 \textcolor{black!55}{\tiny read at 4k, base finishes 6\%}};
\fill[reliaink] (0.4255,-0.12) rectangle (0.4298,0.12);
\node[anchor=west,font=\scriptsize,inner sep=1pt] at (0.4398,0.00) {$+0.002$};
\node[anchor=east,font=\scriptsize,inner sep=1pt] at (-0.02,-0.30) {ProRL v2 \textcolor{black!55}{\tiny read at 8k, base finishes 20\%}};
\fill[finishink] (0.4255,-0.42) rectangle (0.9085,-0.18);
\fill[reliaink] (0.3085,-0.42) rectangle (0.4255,-0.18);
\fill[pattern=north east lines,pattern color=reachink] (0.3064,-0.42) rectangle (0.3085,-0.18);
\node[anchor=west,font=\scriptsize,inner sep=1pt] at (0.9185,-0.30) {$+0.171$};
\node[anchor=east,font=\scriptsize,inner sep=1pt] at (-0.02,-0.60) {e3 \textcolor{black!55}{\tiny read at 16k, base finishes 43\%}};
\fill[finishink] (0.4255,-0.72) rectangle (0.6362,-0.48);
\fill[reliaink] (0.3213,-0.72) rectangle (0.4255,-0.48);
\fill[pattern=north east lines,pattern color=reachink] (0.3170,-0.72) rectangle (0.3213,-0.48);
\node[anchor=west,font=\scriptsize,inner sep=1pt] at (0.6462,-0.60) {$+0.048$};
\node[anchor=east,font=\scriptsize,inner sep=1pt] at (-0.02,-0.90) {JustRL-D \textcolor{black!55}{\tiny read at 16k, base finishes 49\%}};
\fill[finishink] (0.4255,-1.02) rectangle (0.7234,-0.78);
\fill[reliaink] (0.4170,-1.02) rectangle (0.4255,-0.78);
\node[anchor=west,font=\scriptsize,inner sep=1pt] at (0.7334,-0.90) {$+0.136$};
\node[anchor=east,font=\scriptsize,inner sep=1pt] at (-0.02,-1.20) {JustRL-N \textcolor{black!55}{\tiny read at 16k, base finishes 66\%}};
\fill[finishink] (0.4255,-1.32) rectangle (0.4319,-1.08);
\fill[reliaink] (0.4319,-1.32) rectangle (0.5830,-1.08);
\fill[pattern=north east lines,pattern color=reachink] (0.5830,-1.32) rectangle (0.5894,-1.08);
\node[anchor=west,font=\scriptsize,inner sep=1pt] at (0.5994,-1.20) {$+0.077$};
\node[anchor=east,font=\scriptsize,inner sep=1pt] at (-0.02,-1.50) {AReaL \textcolor{black!55}{\tiny read at 24k, base finishes 74\%}};
\fill[finishink] (0.4255,-1.62) rectangle (0.4830,-1.38);
\fill[reliaink] (0.4830,-1.62) rectangle (0.5255,-1.38);
\fill[pattern=north east lines,pattern color=reachink] (0.5255,-1.62) rectangle (0.5340,-1.38);
\node[anchor=west,font=\scriptsize,inner sep=1pt] at (0.5440,-1.50) {$+0.051$};
\node[anchor=east,font=\scriptsize,inner sep=1pt] at (-0.02,-1.80) {DeepMath \textcolor{black!55}{\tiny read at 24k, base finishes 74\%}};
\fill[finishink] (0.4255,-1.92) rectangle (0.5745,-1.68);
\fill[reliaink] (0.5745,-1.92) rectangle (0.5979,-1.68);
\fill[pattern=north east lines,pattern color=reachink] (0.5979,-1.92) rectangle (0.6085,-1.68);
\node[anchor=west,font=\scriptsize,inner sep=1pt] at (0.6185,-1.80) {$+0.086$};
\node[anchor=east,font=\scriptsize,inner sep=1pt] at (-0.02,-2.10) {DeepScaleR \textcolor{black!55}{\tiny read at 24k, base finishes 74\%}};
\fill[finishink] (0.4255,-2.22) rectangle (0.5745,-1.98);
\fill[reliaink] (0.5745,-2.22) rectangle (0.5979,-1.98);
\node[anchor=west,font=\scriptsize,inner sep=1pt] at (0.6079,-2.10) {$+0.081$};
\node[anchor=east,font=\scriptsize,inner sep=1pt] at (-0.02,-2.40) {RLinf \textcolor{black!55}{\tiny read at 28k, base finishes 80\%}};
\fill[finishink] (0.4255,-2.52) rectangle (0.5468,-2.28);
\fill[reliaink] (0.5468,-2.52) rectangle (0.6426,-2.28);
\fill[pattern=north east lines,pattern color=reachink] (0.6426,-2.52) rectangle (0.6574,-2.28);
\node[anchor=west,font=\scriptsize,inner sep=1pt] at (0.6674,-2.40) {$+0.109$};
\node[anchor=east,font=\scriptsize,inner sep=1pt] at (-0.02,-2.70) {STILL-3 \textcolor{black!55}{\tiny read at 28k, base finishes 80\%}};
\fill[finishink] (0.4255,-2.82) rectangle (0.4809,-2.58);
\fill[reliaink] (0.4128,-2.82) rectangle (0.4255,-2.58);
\fill[pattern=north east lines,pattern color=reachink] (0.4809,-2.82) rectangle (0.4979,-2.58);
\node[anchor=west,font=\scriptsize,inner sep=1pt] at (0.5079,-2.70) {$+0.028$};
\node[anchor=east,font=\scriptsize,inner sep=1pt] at (-0.02,-3.00) {Skywork-32B \textcolor{black!55}{\tiny read at 24k, base finishes 88\%}};
\fill[finishink] (0.3894,-3.12) rectangle (0.4255,-2.88);
\fill[reliaink] (0.4255,-3.12) rectangle (0.6915,-2.88);
\fill[pattern=north east lines,pattern color=reachink] (0.6915,-3.12) rectangle (0.7085,-2.88);
\node[anchor=west,font=\scriptsize,inner sep=1pt] at (0.7185,-3.00) {$+0.116$};
\node[anchor=east,font=\scriptsize,inner sep=1pt] at (-0.02,-3.30) {AceReason \textcolor{black!55}{\tiny read at 32k, base finishes 91\%}};
\fill[finishink] (0.4255,-3.42) rectangle (0.5319,-3.18);
\fill[reliaink] (0.5319,-3.42) rectangle (0.6617,-3.18);
\fill[pattern=north east lines,pattern color=reachink] (0.6617,-3.42) rectangle (0.6766,-3.18);
\node[anchor=west,font=\scriptsize,inner sep=1pt] at (0.6866,-3.30) {$+0.118$};
\node[anchor=east,font=\scriptsize,inner sep=1pt] at (-0.02,-3.60) {Skywork-7B \textcolor{black!55}{\tiny read at 32k, base finishes 91\%}};
\fill[finishink] (0.3191,-3.72) rectangle (0.4255,-3.48);
\fill[reliaink] (0.4255,-3.72) rectangle (0.6745,-3.48);
\fill[pattern=north east lines,pattern color=reachink] (0.6745,-3.72) rectangle (0.6851,-3.48);
\node[anchor=west,font=\scriptsize,inner sep=1pt] at (0.6951,-3.60) {$+0.072$};
\node[anchor=east,font=\scriptsize,inner sep=1pt] at (-0.02,-3.90) {Skywork-Math \textcolor{black!55}{\tiny read at 32k, base finishes 91\%}};
\fill[finishink] (0.4255,-4.02) rectangle (0.4468,-3.78);
\fill[reliaink] (0.4468,-4.02) rectangle (0.6255,-3.78);
\fill[pattern=north east lines,pattern color=reachink] (0.6255,-4.02) rectangle (0.6362,-3.78);
\node[anchor=west,font=\scriptsize,inner sep=1pt] at (0.6462,-3.90) {$+0.099$};
\node[anchor=east,font=\scriptsize,inner sep=1pt] at (-0.02,-4.20) {DeepMath-Omn \textcolor{black!55}{\tiny read at 24k, base finishes 93\%}};
\fill[finishink] (0.4191,-4.32) rectangle (0.4255,-4.08);
\fill[reliaink] (0.4255,-4.32) rectangle (0.5298,-4.08);
\fill[pattern=north east lines,pattern color=reachink] (0.4149,-4.32) rectangle (0.4191,-4.08);
\node[anchor=west,font=\scriptsize,inner sep=1pt] at (0.5398,-4.20) {$+0.044$};
\node[anchor=east,font=\scriptsize,inner sep=1pt] at (-0.02,-4.50) {KnowRL \textcolor{black!55}{\tiny read at 24k, base finishes 93\%}};
\fill[finishink] (0.3106,-4.62) rectangle (0.4255,-4.38);
\fill[reliaink] (0.4255,-4.62) rectangle (0.7447,-4.38);
\fill[pattern=north east lines,pattern color=reachink] (0.7447,-4.62) rectangle (0.7702,-4.38);
\node[anchor=west,font=\scriptsize,inner sep=1pt] at (0.7802,-4.50) {$+0.108$};
\node[anchor=east,font=\scriptsize,inner sep=1pt] at (-0.02,-4.80) {QuestA \textcolor{black!55}{\tiny read at 24k, base finishes 93\%}};
\fill[finishink] (0.0617,-4.92) rectangle (0.4255,-4.68);
\fill[reliaink] (0.4255,-4.92) rectangle (0.8277,-4.68);
\fill[pattern=north east lines,pattern color=reachink] (0.8277,-4.92) rectangle (0.8468,-4.68);
\node[anchor=west,font=\scriptsize,inner sep=1pt] at (0.8568,-4.80) {$+0.027$};
\node[anchor=east,font=\scriptsize,inner sep=1pt] at (-0.02,-5.10) {AceReason-14B \textcolor{black!55}{\tiny read at 32k, base finishes 94\%}};
\fill[finishink] (0.4106,-5.22) rectangle (0.4255,-4.98);
\fill[reliaink] (0.4255,-5.22) rectangle (0.6830,-4.98);
\fill[pattern=north east lines,pattern color=reachink] (0.6830,-5.22) rectangle (0.7277,-4.98);
\node[anchor=west,font=\scriptsize,inner sep=1pt] at (0.7377,-5.10) {$+0.135$};
\node[anchor=east,font=\scriptsize,inner sep=1pt] at (-0.02,-5.40) {Polaris-1.7B \textcolor{black!55}{\tiny read at 48k, base finishes 98\%}};
\fill[finishink] (0.1766,-5.52) rectangle (0.4255,-5.28);
\fill[reliaink] (0.4255,-5.52) rectangle (0.7809,-5.28);
\fill[pattern=north east lines,pattern color=reachink] (0.7809,-5.52) rectangle (0.8064,-5.28);
\node[anchor=west,font=\scriptsize,inner sep=1pt] at (0.8164,-5.40) {$+0.062$};
\node[anchor=east,font=\scriptsize,inner sep=1pt] at (-0.02,-5.70) {Polaris-4B \textcolor{black!55}{\tiny read at 52k, base finishes 98\%}};
\fill[finishink] (0.2021,-5.82) rectangle (0.4255,-5.58);
\fill[reliaink] (0.4255,-5.82) rectangle (0.7213,-5.58);
\fill[pattern=north east lines,pattern color=reachink] (0.7213,-5.82) rectangle (0.7511,-5.58);
\node[anchor=west,font=\scriptsize,inner sep=1pt] at (0.7611,-5.70) {$+0.048$};
\draw[black!60,thin] (0.4255,0.30) -- (0.4255,-6.10);
\draw[black!50,thin] (0,-6.10) -- (1,-6.10);
\foreach \v/\l in {-0.1/$-0.1$,0/$0$,0.1/$0.1$,0.2/$0.2$} \node[anchor=north,font=\scriptsize] at ({(\v+0.20)*2.1277},-6.12) {\l};
\node[anchor=north,font=\scriptsize] at (0.5,-6.40) {change in pass@1 over the base, competition math};
\end{tikzpicture}}\\[2pt]
{\scriptsize \tikz\fill[finishink] (0,0) rectangle (0.3,0.18); finishing $F$\quad \tikz\fill[reliaink] (0,0) rectangle (0.3,0.18); reliability $P$\quad \tikz\fill[pattern=north east lines,pattern color=reachink] (0,0) rectangle (0.3,0.18); reach, net of its mirror}
\caption{What outcome RL buys at each recipe's own training cap on competition math, the recipes' own domain. Rows are split as in Figure~\ref{fig:recipes}. Parts left of zero read as in Figure~\ref{fig:overview}. Rows are sorted by how often the base finishes at the cap, least first. }
\label{fig:mathrecipes}
\end{figure}
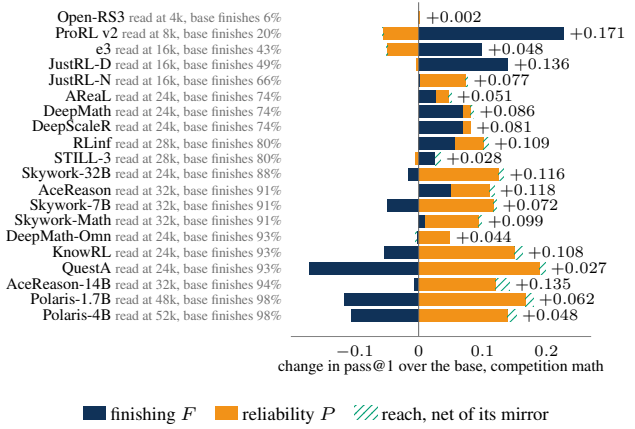

\subsection{Metrics}
\label{sec:metrics}
We split a pass@$k$ gain under a token budget $B$ into three parts: \emph{finishing} $F$, answering in time; \emph{reliability} $P$, answering right once finished; and \emph{reach}, reach@$k(B)$, solving problems the base leaves unsolved in $k$ samples. A \emph{finished} derivation stops within the budget.
With $k$ samples per problem at budget $B$, pass@$k(B)$ is the share of problems with at least one correct finished sample. The finishing rate fin@$k(B)$ is the share with at least one finished sample. The reliability rate pass@$k(B\,|\,\text{fin})$ is the share of problems with a finished sample that also have a correct one, so pass@$k(B)$ factors as
\begin{equation}
\text{pass@}k(B)=\text{fin@}k(B)\times\text{pass@}k(B\,|\,\text{fin}),
\label{eq:finrel}
\end{equation}
and a change splits exactly into a finishing term $F$ and a reliability term $P$ (reach is split off below):
\begin{align}
F&=\Delta\text{fin@}k(B)\times\text{pass@}k_{\mathrm{RL}}(B\,|\,\text{fin}),\label{eq:F}\\
P&=\text{fin@}k_{\mathrm{base}}(B)\times\Delta\text{pass@}k(B\,|\,\text{fin}).\label{eq:P}
\end{align}
Here $F$ takes the cross term, $\Delta\text{fin@}k(B)\times\Delta\text{pass@}k(B\,|\,\text{fin})$. Giving it to $P$, or half to each, gives the split's other two orders.

\emph{Reach}, reach@$k(B)$, asks the question of \citet{yue2025} at a given $k$: how much does RL score on problems its base leaves unsolved in $k$ samples? A model leaves a problem \emph{unsolved} when it finishes all $k$ samples and none is correct. Requiring all $k$ to finish rules out problems lost only to the budget. Out of $n$ problems, let $U$ be those the base leaves unsolved and $U'$ those the RL model leaves unsolved. With $k$ samples, either model can miss a problem it solves with low probability, so winning a problem alone can be chance. Reach therefore subtracts the \emph{mirror}: what the base scores on the problems the RL model leaves unsolved.
\begin{equation}
\begin{split}
\text{reach@}k(B)=\frac{1}{n}\Big(&\sum_{q\in U}\text{pass@1}_{\mathrm{RL}}(q;B)\\
&-\sum_{q\in U'}\text{pass@1}_{\mathrm{base}}(q;B)\Big).
\end{split}
\label{eq:reach}
\end{equation}
$k$ only decides which problems count as unsolved. Each model is scored on them by pass@1, so reach is in pass@1 units. $F$ and $P$ are computed at $k=1$ on the problems outside $U$, weighted by their share of the $n$. Reach subtracts the mirror and $P$ adds it back, so nothing counts twice. The parts sum exactly: $\Delta\text{pass@1}(B)=F+P+\text{reach@}k(B)$.
\emph{Robust} reach ignores problems solved once by chance. Robust reach keeps only unsolved problems the other model solves in at least $k/2$ of its $k$ samples, the sets $U_{\geq k/2}\subseteq U$ and $U'_{\geq k/2}\subseteq U'$:
\begin{equation}
\begin{split}
\text{reach}_{\geq k/2}\text{@}k(B)=\frac{1}{n}\Big(&\sum_{q\in U_{\geq k/2}}\text{pass@1}_{\mathrm{RL}}(q;B)\\
&-\sum_{q\in U'_{\geq k/2}}\text{pass@1}_{\mathrm{base}}(q;B)\Big).
\end{split}
\label{eq:robustreach}
\end{equation}
\emph{Matched} reliability is reliability measured only on problems both models finish. For the released pairs, both must finish at least seven of eight samples.

\subsection{Protocol for Reading the Released Pairs}
\label{sec:released}
We read twenty released outcome-RL models, each against the thinking base it was trained from (Table~\ref{tab:recipes}). Eighteen are of up to $7$B. The other two are a $14$B and a $32$B pair. Each RL model and its base are sampled the same way, with one prompt and one answer checker, the matcher.  Each model writes $k=8$ answers per problem at temperature $T=0.6$. Budgets: $4$k, $8$k and $16$k; the recipe's training cap; and the next budget above the cap on a fixed grid up to $64$k. The split is read at $k=1$, per sample. \finB{} is the share of derivations that finish within $B$. \relB{} is pass@1 on finished derivations. Reach is read at $k=8$, as reach@8 and reach$_{\geq4}$@8 ($k=32$ in Appendix~\ref{sec:reach32}). We \emph{registered} predictions for ten pairs at $4$k and $8$k, committing them before any of their outputs existed. Reads at the cap and the next budget above it were not registered.

\section{Finishing, Reliability, Rarely Reach}
\label{sec:what}
\subsection{Finishing}
\label{sec:termination}

\begin{table}[!t]
\centering\footnotesize
\setlength{\tabcolsep}{3pt}
\begin{tabularx}{\columnwidth}{>{\raggedright\arraybackslash}Xl}
\toprule
\multicolumn{2}{l}{\emph{base derivations that finish, physics / MATH-500}} \\
\quad at $4$k and $8$k, range over bases & $12$--$65\%$ / $56$--$90\%$ \\
\quad at $16$k & $82$--$93\%$ / $92$--$97\%$ \\
\multicolumn{2}{l}{\emph{finishing's share of the change, $4$k and $8$k}} \\
\quad both sets & $82\%$ [$77$, $86$] \\
\quad physics / MATH-500 & $71\%$ / $92\%$ \\
RL's change in length vs.\ in finishing & $\rho=-0.93$ \\
Open-Reasoner-Zero-7B, finishing at $2$k & $0.98\to0.65$ \\
\multicolumn{2}{l}{\emph{base score, tight budget $\to$ $64$k}} \\
\quad problems picked at $4$k (RL $0.68$) & $0.44\to0.80$ \\
\quad problems picked at $8$k (RL $0.65$) & $0.38\to0.59$ \\
\bottomrule
\end{tabularx}
\caption{Finishing at tight budgets. Arrows read base to RL, except in the last block. Share $=\sum|F|/(\sum|F|+\sum|P|)$, with a $95\%$ interval. $\rho$: Spearman's correlation over recipes.}
\label{tab:finishing}
\end{table}

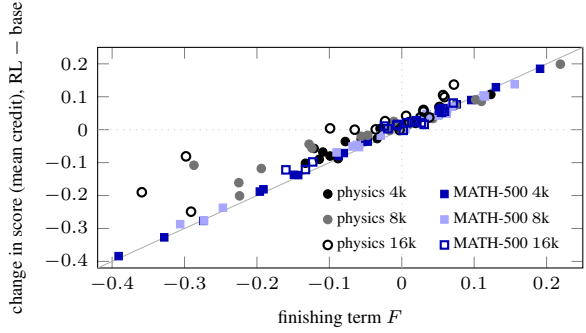
\begin{figure}[t]
\centering
\begin{tikzpicture}
\begin{axis}[width=8.0cm,height=4.5cm,xmin=-0.42,xmax=0.25,ymin=-0.42,ymax=0.25,xlabel={finishing term $F$},ylabel={change in score (mean credit), RL $-$ base},
  tick label style={font=\scriptsize},label style={font=\scriptsize},xtick={-0.4,-0.3,-0.2,-0.1,0,0.1,0.2},ytick={-0.4,-0.3,-0.2,-0.1,0,0.1,0.2},
  legend style={font=\tiny,at={(0.98,0.02)},anchor=south east,draw=none,fill=none,legend columns=2,/tikz/every even column/.append style={column sep=0.25cm}},legend cell align=left]
\addplot[gray!60,thin,forget plot] coordinates {(-0.42,-0.42) (0.25,0.25)};
\addplot[gray!40,dotted,thin,forget plot] coordinates {(-0.42,0) (0.25,0)};
\addplot[gray!40,dotted,thin,forget plot] coordinates {(0,-0.42) (0,0.25)};
\addplot[only marks,mark=*,mark size=1.6pt,black] coordinates {(0.041,0.037) (-0.030,0.003) (0.001,0.003) (0.055,0.049) (-0.133,-0.102) (-0.114,-0.090) (-0.109,-0.068) (-0.088,-0.088) (-0.003,0.006) (-0.056,-0.029) (0.123,0.107) (0.002,0.007) (-0.034,-0.027) (-0.121,-0.057) (-0.078,-0.036) (-0.048,-0.023) (0.019,0.019) (-0.099,-0.080)};
\addplot[only marks,mark=square*,mark size=1.4pt,blue!70!black] coordinates {(0.130,0.129) (0.053,0.066) (0.003,0.002) (0.096,0.090) (-0.149,-0.137) (-0.087,-0.080) (-0.080,-0.071) (-0.274,-0.277) (-0.047,-0.036) (-0.050,-0.036) (0.191,0.185) (0.035,0.038) (-0.196,-0.188) (-0.391,-0.384) (-0.191,-0.181) (-0.143,-0.138) (0.076,0.077) (-0.328,-0.327)};
\addplot[only marks,mark=*,mark size=1.6pt,black!55] coordinates {(0.110,0.086) (0.219,0.199) (0.001,-0.002) (0.062,0.056) (-0.194,-0.118) (-0.047,-0.017) (-0.124,-0.056) (-0.224,-0.201) (0.074,0.074) (-0.011,0.025) (0.102,0.091) (0.010,0.016) (-0.017,-0.002) (-0.287,-0.108) (-0.128,-0.044) (-0.056,-0.020) (0.043,0.034) (-0.225,-0.161)};
\addplot[only marks,mark=square*,mark size=1.4pt,blue!35] coordinates {(0.115,0.101) (0.156,0.138) (0.005,0.002) (0.062,0.049) (-0.068,-0.051) (0.025,0.036) (-0.029,-0.018) (-0.273,-0.276) (0.069,0.070) (0.034,0.048) (0.113,0.104) (0.037,0.036) (-0.059,-0.053) (-0.306,-0.287) (-0.090,-0.069) (-0.062,-0.046) (0.050,0.043) (-0.247,-0.237)};
\addplot[only marks,mark=o,mark size=1.6pt,black,thick] coordinates {(0.052,0.069) (0.072,0.137) (-0.003,-0.001) (0.026,0.038) (-0.065,0.000) (0.030,0.055) (-0.023,0.026) (-0.291,-0.249) (0.059,0.099) (0.057,0.105) (0.030,0.060) (0.010,0.023) (0.006,0.042) (-0.298,-0.081) (-0.099,0.004) (-0.036,0.001) (0.038,0.037) (-0.359,-0.190)};
\addplot[only marks,mark=square,mark size=1.4pt,blue!70!black,thick] coordinates {(0.059,0.056) (0.071,0.081) (0.002,-0.002) (0.027,0.021) (-0.019,0.003) (0.014,0.023) (-0.006,0.016) (-0.133,-0.121) (0.057,0.064) (0.055,0.053) (0.052,0.059) (0.020,0.027) (0.007,0.021) (-0.160,-0.122) (-0.024,0.011) (-0.007,0.013) (0.031,0.016) (-0.123,-0.098)};
\legend{physics $4$k,MATH-500 $4$k,physics $8$k,MATH-500 $8$k,physics $16$k,MATH-500 $16$k}
\end{axis}
\end{tikzpicture}
\caption{Change in score against the finishing term $F$, with Spearman $\rho=0.97$ over all $108$ points. Each point is one pair on one set at one budget. A point on the diagonal is a change carried by finishing alone. A point's vertical distance from the diagonal is the rest of the change: reliability, reach and partial credit. The correlation is $0.99$ at $4$k and $8$k and $0.92$ at $16$k.}
\label{fig:pairs}
\end{figure}

\textbf{When the base often hits the budget before answering, finishing carries most of the change.} At $4$k and $8$k the bases leave many derivations unfinished. Finishing is then the larger part of the change in every recipe up to $7$B except Open-RS3, whose score moves by at most $0.004$ at these budgets (Figure~\ref{fig:recipes}). The reward explains why. An unfinished derivation scores zero whatever it contains, so at a tight budget the first thing group-relative credit can tell apart is whether a sample finished. The finishing share is larger on MATH-500 than on physics, because a finished base derivation on MATH-500 is already usually right. Figure~\ref{fig:pairs} shows each cell's change tracking its finishing term.

\textbf{Outcome RL moves derivation length toward what earns most reward within its cap.} The effect on finishing therefore depends on where the base starts. At a short cap, RL teaches a thinking base to fit its derivations inside the cap, as ProRL v2 does at $8$k. At a long cap, RL teaches longer derivations. The RL model then finishes less often, as Polaris-1.7B does even at its own cap. Across recipes, the change in derivation length tracks the change in finishing (Table~\ref{tab:finishing}). RL lengthens derivations on a short-answer base too, and the RL model then finishes less at a tight budget (Open-Reasoner-Zero-7B \citep{orz2025}; Table~\ref{tab:finishing}).

\textbf{Finishing is not hidden knowledge.} A base that cannot finish within $B$ tokens might not know how to derive the answer. Two results argue against this reading. First, the extra tokens go to hesitation, not derivation. In our controlled run (\S\ref{sec:install}), RL halves the backtracking words (``wait'', ``actually'') per derivation. Reliability on problems finished both before and after training does not move. Second, given more tokens, the base solves most of the problems on which RL, at a tight budget, finishes more often and scores higher. We pick those problems with four of the eight samples, then score them on the other four and on a separate $64$k base run. At $64$k the base passes the RL model on the problems picked at $4$k and closes most of the gap on those picked at $8$k (Table~\ref{tab:finishing}). The gap the base does not close counts as reliability, not finishing. Finishing and knowledge are not fully separable, since committing to the right approach is a skill. Our claim is narrower: RL adds few steps the base does not already write (\S\ref{sec:reach}). Finishing still pays at a tight deployment budget, as ProRL v2 at $8$k shows.

\subsection{Reliability, Where the Base Already Finishes}
\label{sec:reliability}

\begin{table}[!t]
\centering\footnotesize
\setlength{\tabcolsep}{3pt}
\begin{tabularx}{\columnwidth}{>{\raggedright\arraybackslash}Xrr}
\toprule
physics at the cap, mean per recipe & $F$ & $P$ \\
\midrule
four cap-bound recipes & $+0.078$ & $+0.002$ \\
\quad matched split & $+0.058$ & $+0.021$ \\
\quad Open-RS3, total change$^\ast$ & \multicolumn{2}{c}{$+0.003$} \\
sixteen recipes that are not cap-bound & $+0.002$ & $+0.032$ \\
all twenty recipes & $+0.017$ & $+0.026$ \\
\multicolumn{3}{l}{\emph{competition math, matched split}} \\
\quad base finishes $<90\%$ & $+0.045$ & $+0.035$ \\
\quad base finishes $\geq90\%$ & $-0.011$ & $+0.081$ \\
\midrule
\multicolumn{3}{l}{\emph{sixteen recipes that are not cap-bound}} \\
\quad reliability rises & \multicolumn{2}{r}{$16$ of $16$} \\
\quad reliability is larger by size & \multicolumn{2}{r}{$14$ of $16$} \\
\quad matched reliability rises (clear of $0$) & \multicolumn{2}{r}{$16$ ($12$) of $16$} \\
\quad bases with $P>F$ / larger by size & \multicolumn{2}{r}{$7$ / $6$ of $7$} \\
\multicolumn{3}{l}{\emph{all twenty recipes}} \\
\quad mean reach & \multicolumn{2}{r}{$+0.003$} \\
\quad finishing share vs.\ unfinished at the cap & \multicolumn{2}{r}{$\rho=0.59$} \\
\quad reliability rises at the next budget & \multicolumn{2}{r}{$18$ of $20$} \\
\quad finishing rises at the next budget & \multicolumn{2}{r}{$11$ of $20$} \\
\midrule
\multicolumn{3}{l}{\emph{step check: RL's decisive steps the base lacks}} \\
\quad base right $1$--$4$ of $8$, $16$k & \multicolumn{2}{r}{$0$ of $26$} \\
\quad base right $5$--$7$, $13$ caps & \multicolumn{2}{r}{$0$ of $26$} \\
\quad base unfinished, never right, $16$ caps & \multicolumn{2}{r}{$1$ of $57$} \\
\quad reverse, base's steps RL lacks, $16$k & \multicolumn{2}{r}{$0$ of $19$} \\
\bottomrule
\end{tabularx}
\caption{Finishing and reliability by recipe group, and the step check, on physics at the cap unless noted. $\rho$: Spearman's correlation over recipes. At the next budget, a rise is a gain above $0.005$. $^\ast$With $95\%$ interval [$-0.007$, $+0.014$].}
\label{tab:reliability}
\end{table}

\textbf{Whether the base still hits the training cap before answering decides which part grows.} RL can buy finishing only where the base fails to finish (Figure~\ref{fig:recipes}). A recipe is \emph{cap-bound} when its base still often hits its cap before answering. Four recipes are cap-bound: ProRL v2, Open-RS3, e3 and JustRL-DeepSeek. For ProRL v2 and e3 on physics, finishing is the whole gain (Table~\ref{tab:capcells}). JustRL-DeepSeek gains about a third by reliability, half under the matched split. Open-RS3's score does not measurably change at its cap. The split of Eqs.~\eqref{eq:F}--\eqref{eq:P} can hide a reliability gain, because the problems RL newly finishes are harder and pull its reliability down. \emph{Matched} reliability, measured only on problems both models finish, avoids that pull (\S\ref{sec:metrics}). With matched reliability, the four recipes gain some reliability as well. Across all twenty recipes, the more the base leaves unfinished at the cap, the larger the finishing share.

\textbf{Where the base already finishes, reliability rises in every recipe.} The sixteen recipes that are not cap-bound train at caps where the base already finishes $93$--$100\%$ of its derivations. In all sixteen, reliability rises on physics. By size, reliability is the larger part in fourteen of them (Figure~\ref{fig:recipes}). Averaged by base, reliability exceeds finishing on all seven bases, and is larger by size on six (Table~\ref{tab:reliability}). Finishing moves either way: seven of the sixteen finish less than their base. Math shows the same rise, on MATH-500 and on competition math, a separate set of $276$ recent contest problems (Figure~\ref{fig:mathrecipes}). One recipe goes the other way: e3's reliability falls on MATH-500, although its base finishes there.

\textbf{Reliability still rises on problems both models finish.} Matched reliability still rises in all sixteen recipes, clear of zero in twelve (Table~\ref{tab:capcells}). The Polaris pairs finish less than their bases, and their matched rise is smaller than their plain rise.  On competition math, the matched split narrows the gap between the two parts in both groups of Figure~\ref{fig:overview}, without changing which is larger (Table~\ref{tab:reliability}). A registered prediction that the rise would stay under $0.03$ held with both benchmarks pooled but failed on physics alone (Appendix~\ref{app:pairs}). 

\textbf{On problems where RL is right more often, the step behind its win is one the base already writes.} To test whether RL's reliability gain adds a step the base lacks, we drew $40$ problems at random where the base finishes all eight samples at $16$k but is right in only one to four, and the RL model is right more often. A language-model reader, blind to which model wrote which, named the step that makes the RL derivation right and the base derivation wrong. A second reader looked for that step in the base's other samples. On $12$ of the $40$ problems, the two derivations differ only in answer form or grading. None of the $26$ steps judged decisive on the other $28$ is missing from the base (Table~\ref{tab:reliability}). The base's own wrong samples state the step too, but on $17$ of the $26$ they do not carry it through. The reverse search, for the base's steps in the RL model's samples, finds no step missing either. At the caps of $13$ of the $16$ recipes that are not cap-bound, the check finds no RL step missing where the base is right five to seven times. RL uses, more reliably, a step its base already has.

\textbf{Across all twenty recipes, reliability is the part of the gain that rises consistently.} At the cap, reliability is the largest part of the mean physics gain (Table~\ref{tab:reliability}). Across recipes, reliability and reach are clear of zero. Finishing is not, because it moves either way. Resampling the seven bases instead of problems widens the intervals of the two shares until they overlap (Appendix~\ref{app:pairs}). At the next budget above the cap, reliability rises in more recipes than finishing does (Table~\ref{tab:reliability}). Our reads cannot explain why some recipes gain more reliability than others. The recipes differ in data, steps and algorithm at once.

\subsection{Rarely Reach}
\label{sec:reach}

\begin{table}[t]
\centering\footnotesize
\setlength{\tabcolsep}{4pt}
\begin{tabular}{>{\raggedright\arraybackslash}p{4.9cm}cc}
\toprule
physics & RL & base \\
\midrule
solves what the other solves ($16$k) & $83\%$ & $85\%$ \\
lone wins, of $5{,}400$ pair--problems & $392$ & $457$ \\
lone wins, other wrong in all $8$ & $128$ & $153$ \\
solves other's all-wrong, at cap & $14\%$ & $11\%$ \\
\quad with $24$ more samples & $23\%$ & $27\%$ \\
\addlinespace
blind: a step the other lacks ($143$) & $80$ & $71$ \\
upheld by an author, of decided & $10/26$ & $9/19$ \\
found in the other's samples & $49/60$ & $52/62$ \\
\bottomrule
\end{tabular}
\caption{Lone wins and new steps, for the RL model and its base. Rows one to three are at $16$k, rows four and five at the caps of the eighteen smaller recipes. Row one is the share of the other model's solved problems this model also solves. The last three rows read $143$ derivation pairs at $16$k ($23$ on math), the RL one right and the base one wrong. They count steps one has and the other lacks, those an author upheld in a random sample of $50$ pairs, and, on physics, those found in four other samples of the model lacking them.}
\label{tab:sides}
\end{table}

\begin{table}[!t]
\centering\footnotesize
\setlength{\tabcolsep}{3pt}
\begin{tabularx}{\columnwidth}{>{\raggedright\arraybackslash}Xl}
\toprule
\multicolumn{2}{l}{\emph{RL lone wins at $16$k on four bases}} \\
\quad checked & $121$ \\
\quad solved by the base, $56$ more samples & $87$ \\
\quad no longer lone wins when regenerated & $22$ \\
\quad differ only in answer form & $9$ \\
\quad real physics & $3$ \\
RL lone wins at the caps: real physics & $2$ of $23$ \\
\multicolumn{2}{l}{\emph{net reach}} \\
\quad harder physics set ($671$ problems) & $-0.006$ to $0.018$ \\
\quad competition math ($276$ problems) & $-0.002$ to $0.021$ \\
\quad MATH-500, $18$ recipes & $-0.003$ to $0.005$ \\
\quad physics, $1.5$B to $4$B recipes & $-0.007$ to $+0.007$ \\
\quad physics, $7$B and up & $+0.009$ to $+0.013$ \\
\quad $32$B recipe, matcher / answer judge & $0.013$ / $0.004$ \\
\multicolumn{2}{l}{\emph{base never right, some samples unfinished$^\ddagger$}} \\
\quad gain share, not cap-bound / cap-bound & $10\%$ / $34\%$ \\
\quad of RL's wins there, base solves at $64$k & $8\%$ / $44\%$ \\
\quad net reach if unsolved, not cap-bound & $\leq0.019$ \\
\quad ProRL v2's net reach if unsolved$^\P$ & $0.067$ \\
robust wins$^\S$: QuestA/KnowRL/ProRL v2 & $1$ / $3$ / $3$ \\
\quad their bases & $0$ / $4$ / $0$ \\
training steps vs.\ robust reach & $\rho=0.13$ \\
$32$B at its cap: gain / reliability & $+0.052$ / $+0.047$ \\
\bottomrule
\end{tabularx}
\caption{Lone wins checked, and reach on other sets and at other sizes, physics at each recipe's cap unless noted. $^\S$With MATH-500. The $7$B-and-up range covers four of five recipes. $\rho$: Spearman's correlation over the $16$ recipes with stated steps. $^\ddagger$The $16$ regenerated pairs, three of them cap-bound (not e3). $^\P$JustRL-DeepSeek's would be $0.013$ and Open-RS3's $-0.001$.}
\label{tab:reachfollow}
\end{table}

\textbf{Of what RL solves on physics, its base already solves $85\%$.} Problems only one model solves, \emph{lone wins}, are about as common in each direction (Table~\ref{tab:sides}). In two thirds of them, the losing model left at least one sample unfinished. The rest, where the losing model finished all eight and got all eight wrong, are the candidates for new knowledge.

\textbf{A lone win can be chance.} Eight samples can miss a problem a model solves with low probability. Given $24$ more samples, each model solves a similar share of the problems the other got wrong in all eight (Table~\ref{tab:sides}).

\textbf{Of $144$ lone wins we checked, $5$ are real physics.} A lone win is real physics if the base's answers are wrong in substance in every sample we have, not only in form such as an option letter. On four bases at $16$k, the base solves $87$ of RL's $121$ lone wins in $56$ more samples (Table~\ref{tab:reachfollow}). On the other $34$, we read new samples of both models that keep the full text. Three are real physics. Reading $23$ lone wins at the caps, without resampling, finds two more.

\textbf{Reach stays at or below $0.021$ on all four sets.} The bound holds on our physics set, on a harder, unfiltered physics set, on MATH-500 and on competition math (Table~\ref{tab:reachfollow}).

\textbf{A tenth to a third of the gain falls on problems the base never gets right but leaves some samples unfinished.} Eq.~\eqref{eq:reach} needs all eight base samples to finish, so it counts such a problem toward finishing and reliability. At the cap on the $16$ regenerated pairs, these problems carry about a tenth of the gain for the recipes that are not cap-bound and about a third for the cap-bound ones (Table~\ref{tab:reachfollow}). Given $64$k tokens, the bases solve under half of the ones RL gets right. Counted as unsolved, these problems would lift net reach above $0.02$ only for ProRL v2 at its $8$k cap (Table~\ref{tab:reachfollow}). Our definition counts this part as behaviour. On $56$ of the $57$ such problems checked where a step decides, the base's own samples write RL's deciding step (Appendix~\ref{sec:traces}).

\textbf{In a random sample, nine of ten new RL steps a human upheld are already in the base's samples.} On problems an RL model gets right and its base gets wrong, language-model readers compared the winning RL derivation with a failed base one. Read blind, each derivation has a step the other lacks about as often. In a random sample of $50$ pairs, an author upheld under half of these claims on each side (Table~\ref{tab:sides}). Nine of the ten upheld RL steps appear in the base's own samples. A reader then searched four other samples of the model whose derivation lacked the step. On physics, the search finds RL-side steps in the base's samples about as often as the reverse (Table~\ref{tab:sides}).

\textbf{Recipes built to add knowledge solve few new problems.} QuestA puts a partial reference solution in every training prompt. KnowRL adds knowledge-point hints to about four prompts in five. ProRL v2 trains $3{,}000$ steps on data that includes physics. At its cap, each of the three robustly solves at most three problems its base gets wrong in all eight (Table~\ref{tab:reachfollow}). On the harder, unfiltered physics set, QuestA and KnowRL lead the twenty recipes in net reach (Table~\ref{tab:hardcells}). Read one by one, $13$ of their $19$ robust wins there stand, and none is answer form (Appendix~\ref{sec:hardset}). KnowRL trains on QuestA's problems, so the lead may come from those problems rather than the hints. Training length does not predict reach (Table~\ref{tab:reachfollow}). 

\textbf{Larger bases split the same way, and four of the five show more reach.} The three $7$B recipes split like the smaller ones. The $14$B and $32$B recipes train at long caps, so at $4$k and $8$k they finish less often than their bases and score lower. At its cap, the $32$B recipe gains mostly reliability (Table~\ref{tab:reachfollow}). For four of the five recipes from $7$B up, net reach at the cap is higher than for any smaller recipe. Part of this extra reach is answer form, which the matcher rejects more often for larger models. Six of the $32$B recipe's nine robust wins on physics are answer form. A more lenient answer judge removes most of the $32$B recipe's net reach. The judge does not lower the $7$B recipes' reach, so for them the answer-form reading rests only on their robust counts (Table~\ref{tab:robustreach}). Whether real reach grows with scale is open.

\section{A Bounded Probe: Can Supplied Knowledge Be Installed at Small Scale?}
\label{sec:install}\label{sec:controlled}\label{sec:knowrecipes}\label{sec:mechanism}

\begin{table}[!t]
\centering\footnotesize
\setlength{\tabcolsep}{3pt}
\begin{tabularx}{\columnwidth}{>{\raggedright\arraybackslash}p{1.6cm}>{\raggedright\arraybackslash}Xl}
\toprule
channel & in training & held-out vs.\ plain RL \\
\midrule
hint in the prompt & setup claims $55\to92\%$ & $+0.01$ [$-0.01$, $0.03$] \\
teacher derivation & touched problems $12\to59\%$ & $-0.05$ [$-0.08$, $-0.03$] \\
written setup & correct sections $0.29\to0.45$ & $-0.03$ [$-0.05$, $-0.01$] \\
question fine-tune & $99\%$ open with questions & at the base \\
setup reward & the policy gamed its scorer & not measurable \\
\midrule
\multicolumn{2}{l}{\emph{plain RL, the control}} & \emph{value} \\
\multicolumn{2}{l}{plain RL, held-out score} & $0.16\to0.26$ \\
\multicolumn{2}{l}{revisited problems: solve rate} & $0.10\to0.28$ \\
\multicolumn{2}{l}{credited problems: finishing share} & \\
\multicolumn{2}{l}{\quad Eq.~\eqref{eq:F} / even split} & $71\%$ / $50\%$ \\
\multicolumn{3}{l}{\emph{twins, credited minus other}} \\
\multicolumn{2}{l}{\quad planned: mean score} & $+0.096$ [$0.008$, $0.182$] \\
\multicolumn{2}{l}{\quad planned: solved once in eight} & $+0.095$ [$-0.054$, $0.242$] \\
\multicolumn{2}{l}{\quad adjusted for the base$^\dagger$} & $+0.077$ [$-0.004$, $0.160$] \\
\multicolumn{2}{l}{\quad\quad on training problems} & $+0.186$ \\
\multicolumn{2}{l}{\quad minus the fine-tune's excess$^\dagger$} & $+0.061$ [$-0.029$, $0.149$] \\
\multicolumn{2}{l}{\quad both finish $7$ of $8$$^\dagger$} & $-0.003$ [$-0.048$, $0.037$] \\
\bottomrule
\end{tabularx}
\caption{The controlled probe on Qwen3-VL-8B-Thinking. Held-out differences from plain RL are on $87$ hard problems, with $95\%$ intervals. $^\dagger$Chosen after seeing the data.}
\label{tab:probe}
\end{table}

\textbf{At this scale, reward raises reliability only on the problems it lands on. Beyond those problems, the planned test finds a gain whose part beyond finishing is not robust.} To see whether outcome RL installs knowledge it is given, we supply it ourselves, at small scale. We train Qwen3-VL-8B-Thinking \citep{qwen3vl} with group-relative outcome RL (GSPO; \citealp{gspo,grpo}) on physics at a $12$k-token cap, for up to sixty updates, one seed per treated arm and two for the control. Plain RL repeats the result of \S\ref{sec:what}: its held-out gain comes from finishing, with reliability flat (Appendix~\ref{sec:noinstall}). The knowledge supplied is the \emph{setup}: the physical modelling a problem needs before any algebra. The base solves at most $16$ of the $160$ hard training problems at least once in eight samples. A gain on problems the base finishes and still gets wrong would count as new knowledge. Five \emph{channels} supply the setup, from a hint in the prompt to a teacher's full derivation. Each channel changes the model during training (Table~\ref{tab:probe}). None of the four channels we could evaluate is shown to carry over to held-out problems beyond plain RL. The test would detect a gain a third the size of plain RL's. A smaller gain could go undetected, such as the hint's lead over the second control seed (Table~\ref{tab:power}). The fifth channel, a setup reward, taught the policy to game its scorer, so its held-out effect could not be measured.

\textbf{Reward works where it lands.} Training again and again on $60$ hard problems, \emph{revisiting}, raises their solve rate from $0.10$ to $0.28$ (Table~\ref{tab:probe}). Revisiting moves neither the other $293$ hard problems nor the held-out set. On the training problems the control earned reward on, finishing and reliability both rise, so finishing's share of the gain depends on the order of the split (Table~\ref{tab:probe}).

\textbf{Transfer beyond those problems is not shown.} We call a training problem \emph{credited} when the control earned reward on it in updates $31$ to $60$. Each hard training problem has a text-only \emph{twin}: the same physics and method on a new, much easier surface. On the planned mean score, credited problems' twins gain more than other twins, clear of zero (Table~\ref{tab:probe}). The other planned measure, solving a twin at least once in eight samples, includes zero. Over the two control seeds and the three orders of the split, finishing is $26$--$62\%$ of that excess. The base finishes credited twins less often than other twins. The rest is not clear of zero under any order for the first control seed, and only in Eq.~\eqref{eq:F}'s order for the second (Appendix~\ref{sec:twins}). Credited problems also differ in kind: a supervised fine-tune never trained on them shows part of the same excess. Two adjustments added after the full read, for the base and for problem kind, each leave the excess including zero. A last test, chosen after a first look, finds no difference on twins both models finish. A partial transfer cannot be excluded. Even the control's own gains are not shown to transfer beyond finishing. The probe cannot tell whether the channels failed or supplied knowledge does not transfer this way.

\section{Discussion and Limitations}
\label{sec:discussion}

\textbf{The two questions, answered.} On the released models, RL's gain is changed behaviour by our definition, not new knowledge. Where the budget binds, finishing is most of the change ($82\%$ at $4$k and $8$k). Where it no longer binds, the gain is reliability, from steps the base already writes or from answer form. New knowledge, measured as reach, is rare. At our probe's scale, supplied knowledge is not shown to transfer beyond the problems reward lands on. ProRL v2, trained fifty times as long as our probe, holds two of the five real physics wins. A much larger scale may matter.

\textbf{What this means in practice.} Report pass@1 at a budget together with \finB{} at that budget. Otherwise a model that stops sooner will look like one that reasons better. Compare a recipe with its base at or above its training cap. Below the cap, a recipe trained long looks worse than it is. QuestA and the two Polaris recipes score below their bases on physics at $16$k and no longer do at their own caps (Table~\ref{tab:capcells}; $16$k cells released). On MATH-500, QuestA still loses $0.07$ at its cap.

\textbf{Limitations.} The released-pair study and the controlled run are bounded in scale, sampling and grading, and cover thinking bases only. Each limit below names what we checked.

\emph{The probe.} The controlled run has one seed per treated arm, one model family and one domain. The run's held-out set was chosen on base-model samples. The channels ran at up to the control's scale, about two visits per problem. Prolonged training is untested. Supervised distillation at scale, known to transfer \citep{yue2025}, is untested too.

\emph{Sampling.} The released pairs are read with eight samples per problem ($32$ or $64$ on lone-win problems). The budgets are three fixed ones, the cap and the next budget above it, up to $64$k. Only Polaris evaluates higher, at $90$k. The pairs are read at temperature $0.6$, where nine recipes report at $0.7$ to $1.4$. The eight pairs re-read at their authors' temperatures split the same way (Appendix~\ref{sec:temp}). A second generation of $16$ pairs, three of them cap-bound (Table~\ref{tab:regrade2}), moves each term at the cap by at most $0.007$, inside every interval.

\emph{Reach.} Our definition counts as behaviour the tenth to third of the gain on problems the base never gets right but leaves some samples unfinished (\S\ref{sec:reach}). Filtering the physics problems on the controlled run's base caps how much reach they can show. On MATH-500, neither model was resampled on the problems it alone fails. The MATH-500 mirror may therefore hold real losses of coverage. On both physics sets, both models were resampled (Appendices~\ref{sec:reach32} and~\ref{sec:hardset}). Physics is at most about $1\%$ of the recipes' stated RL data but may be in the bases' pretraining. We did not check our problems against any recipe's training set.

\emph{Grading.} The answer matcher is strict, and stricter for larger models. On the $40$ reliability problems checked by language-model readers, $12$ RL wins over the base differ only in answer form or grading. Part of the measured reliability gain therefore reflects which answer forms the matcher accepts. Regraded under the fixed judge protocol of Appendix~\ref{sec:judge-v2}, $3$--$6\%$ of the matcher-rejected physics answers become right for the $1.5$B and $1.7$B models, and $10$--$17\%$ for the $4$B to $32$B models. On the eight pairs with stored texts and the $16$ regenerated pairs, the judge raises \relB{} for base and RL alike. On the $16$ regenerated pairs, the judge lowers the reliability term at the cap by $0.003$ on average, about $8\%$ of it. The judge moves net reach by less than $0.01$ (Table~\ref{tab:regrade2}). Closed models were trace readers (Claude Opus~5.5) and setup grader (Claude Sonnet~5).  The trace readings are by one language model. One author checked the model's claimed new steps. No human checked the step checks' finds. Only the new-step comparison ($143$ pairs) and the step checks were read blind.

\bibliographystyle{icml2026}
\bibliography{refs}

\appendix
\onecolumn
\renewcommand{\floatpagefraction}{0.85}\renewcommand{\textfraction}{0.07}\renewcommand{\topfraction}{0.93}\renewcommand{\bottomfraction}{0.9}\setcounter{totalnumber}{6}\setcounter{topnumber}{4}\setcounter{bottomnumber}{4}

\FloatBarrier
\section{The released pairs}
\label{app:pairs}

\textbf{How the pairs were read.} The base scores zero on $U$, so by Eq.~\eqref{eq:finrel} the change is RL's score on $U$ plus $F+P$ (\S\ref{sec:metrics}). Pooled shares divide $\sum|F|$, $\sum|P|$ and $\max(0,\sum\text{reach@8}(B))$ by their total. $F$ includes the cross term. Counting the cross term as reliability puts the finishing share below one half in only three cells, all physics at $16$k. Counting only full credit moves a cell's change by up to $0.03$.

\textbf{Models and sampling.} Table~\ref{tab:recipes} lists the twenty pairs \citep{deepseekr1,aimo2_2025,qwen3,deepscaler2025,prorl2025,openrs2025,still3_2025,justrl2025,deepmath2025,rlinf2025,areal2025,skyworkor1,acereason2025,questa,knowrl2026,polaris2025,e3_2025}. OpenMath-Nemotron-1.5B is a supervised fine-tune of Qwen2.5-Math-1.5B, without RL. The RL data is mathematics, with code or puzzles in some recipes. Physics is at most about $1\%$ of it in any recipe that states its mix, except ProRL v2. Each model writes eight samples on each of $300$ random physics problems. Physics uses the paper's outcome reward, MATH-500 symbolic equivalence. The two generation protocols agree within $0.01$ on cell means.

\textbf{The base catches up on the problems behind finishing.} Samples $0$--$3$ pick problems where the RL model finishes more and scores higher. Samples $4$--$7$ score them, so selection does not inflate the comparison. The RL model scores $0.68$ at $4$k, the base $0.44$ at $4$k and $0.80$ at $64$k. At $8$k the three scores are $0.65$, $0.38$ and $0.59$. The Qwen3 bases' $64$k read covers physics only.

\textbf{Statistics.} Resampling the seven bases instead widens the at-cap intervals to [$18$, $59$] for finishing and [$38$, $70$] for reliability, which overlap. Reliability's lead at the next budget above the cap stays clear of zero, $34$ points [$19$, $62$]. Pooled over eighteen recipes and both benchmarks with random effects, the matched change is $0.014$ [$0.008$, $0.021$] (between-recipe heterogeneity $I^2=0.73$). On physics alone the matched change is $0.028$ [$0.013$, $0.043$] ($I^2=0.69$). The finished-length ratio is not significantly associated with the reliability term at the cap ($0.35$, $p=0.15$).

\textbf{The shares survive moving the cross term and removing selection.} A matched split, taking the change only on problems both models finish in seven of eight samples, keeps selection out. At the cap on physics, the shares are $45\%$ finishing, $49\%$ reliability and $5\%$ reach, and beyond the cap $28\%$, $62\%$ and $10\%$. The matched split moves them toward reliability, to $37\%$ and $57\%$ at the cap and $19\%$ and $71\%$ beyond, where e3 has no matched read. A symmetric split moves them by a point. For the five recipes whose finishing term is negative at the cap, selection favours reliability: matched reliability is lower for four (Table~\ref{tab:capcells}). Two of the five differences, QuestA's and Skywork-7B's, are within the $0.007$ by which the two generations differ. Under the matched split, the cap-bound recipes' finishing and reliability at the cap are $+0.155$ and $+0.023$ for ProRL v2 (against $+0.203$ and $-0.026$ plain), $+0.047$ and $+0.048$ for JustRL-DeepSeek, $+0.032$ and $+0.009$ for e3, and $0.000$ and $+0.003$ for Open-RS3. The Polaris pairs fall to $+0.015$ and $+0.012$, intervals including zero.

\textbf{Recipes are not independent tests.} The twenty recipes start from seven bases, eight of them from R1-Distill-1.5B. At $4$k and $8$k, finishing is the larger part of the change on both benchmarks together in each of the $17$ smaller recipes whose score changes (one-sided sign test $p=8\times10^{-6}$). Counted by base, five of five gives only $p=0.03$, so we rely on the pooled bootstrap. Where the base finishes, the base-level test is stronger. With the base as a random effect (a one-way random intercept fitted by REML, with a $t$ interval on six degrees of freedom), reliability exceeds finishing at the cap by $+0.033$ [$+0.016$, $+0.049$] over the sixteen recipes that are not cap-bound, and by $+0.030$ [$+0.018$, $+0.043$] at the next budget. Leaving out any one base, the excess at the cap is $+0.030$ to $+0.040$, each clear of zero. The excess is positive on average for each of their seven bases (one-sided sign test $p=0.008$). Over all twenty recipes at the cap it is $+0.014$ [$-0.026$, $+0.053$], because all four cap-bound recipes gain more by finishing. The length association is partly between bases: $-0.78$ within bases ($p=0.006$), $-0.62$ ($p=0.12$) over the eight R1-Distill-1.5B recipes. On physics, AReaL's and DeepMath's finishing-share intervals include one half. We registered no $16$k prediction. At $16$k the first eight pairs gave $71\%$ [$49$, $84$].

\textbf{The registered test.} Predictions for the ten added pairs were committed before their outputs existed. A same-day amendment added four recipes. H1, a finishing share above one half for each added recipe whose score changes at $4$k or $8$k, passes for all ten ($p=0.001$). H1 is weak where every base often hits the budget before answering. H2, a mean matched change below $0.03$ with none above $0.10$, held at $+0.017$ while pooling physics with MATH-500. On MATH-500 the matched base is near ceiling. On physics alone the ten average $+0.042$ and QuestA reaches $+0.108$, outside H2's bounds. Four of five registered $8$k finishing directions held. JustRL-Nemotron finishes less ($-0.05$), not more.

\begin{table}[!htbp]
\centering\scriptsize
\setlength{\tabcolsep}{2.5pt}\renewcommand{\arraystretch}{1.0}
\begin{tabular}{>{\raggedright\arraybackslash}p{2.3cm}>{\raggedright\arraybackslash}p{1.6cm}>{\raggedright\arraybackslash}p{2.8cm}>{\raggedright\arraybackslash}p{2.5cm}>{\raggedright\arraybackslash}p{2.3cm}>{\raggedright\arraybackslash}p{1.5cm}>{\raggedright\arraybackslash}p{0.9cm}>{\raggedright\arraybackslash}p{1.6cm}}
\toprule
model & base & algorithm, scale & length shaping & RL data & training cap & rollout $T$ & authors' evaluation \\
\midrule
DeepScaleR-1.5B-Preview & R1-Distill-1.5B & GRPO with a KL loss; about $1{,}750$ steps & none in the reward; cap raised $8$k$\to16$k$\to24$k & $40$k math & $8$k to $24$k & $0.6$ & $32$k, $T{=}0.6$ \\
ProRL v2 (Nemotron-Research-Reasoning-Qwen-1.5B) & ProRL v1, from R1-Distill-1.5B & REINFORCE++ baseline, clip-higher, dynamic sampling, KL with reference resets; $3{,}000$ steps in all, counting ProRL v1's $2{,}000$ & cosine length penalty, switched on in scheduled cycles & $136$k: math, code, STEM, puzzles, instruction following & $8$k & n.d. & $32$k, $T{=}0.6$ \\
Open-RS3 & R1-Distill-1.5B & GRPO with KL; $50$ steps & cosine length reward & $7$k math & $3{,}584$ & $0.7$ & $32$k, $T{=}0.6$ \\
STILL-3-1.5B-preview & R1-Distill-1.5B & GRPO with KL and reference resets; about $210$ steps (inferred) & none documented & $30$k math & $29$k & $1.0$ & $32$k, $T{=}0.6$ \\
Skywork-OR1-7B & R1-Distill-7B & GRPO without KL, adaptive entropy, rejection sampling; $1{,}320$ steps & none; cap raised $16$k$\to32$k & $105$k math, $14$k code & $16$k to $32$k & $1.0$ & $32$k, $T{=}1.0$ \\
Skywork-OR1-Math-7B & R1-Distill-7B & as above, two updates per step; $2{,}160$ steps & none; cap raised $8$k$\to32$k & Skywork set (mix not stated) & $8$k to $32$k & $1.0$ & $32$k, $T{=}1.0$ \\
AceReason-Nemotron-7B & R1-Distill-7B & on-policy GRPO without KL; math, then code, then math & none; cap raised $8$k$\to32$k & $49$k math, $8.5$k code & $8$k to $32$k & n.d. & $32$k, $T{=}0.6$ \\
Polaris-4B-Preview & Qwen3-4B & GRPO without KL or entropy loss, clip-higher; about $700$ steps & none; cap raised $40$k$\to52$k & $30$k math & $40$k to $52$k & $1.4$ to $1.5$ & $90$k, $T{=}1.4$ \\
\bottomrule
\end{tabular}
\caption{What each released recipe did, from its model card, paper, blog or training scripts (n.d.: not documented). This part lists the eight
recipes read first. The ten added are in the continuation. Rewards are outcome rewards unless a length term is listed. Outcome rewards are $0$ or $1$, except $\pm1$ for STILL-3, JustRL and DeepMath and $\pm5$
for RLinf and AReaL. A truncated response usually scores as a wrong one. A truncated response scores $-2$ under STILL-3, which
docks a missing boxed answer, and under DeepMath's overlong penalty. Some verifiers credit a truncated response if a correct boxed answer appears
before the cut. ``Authors' evaluation'' is the generation budget and temperature its authors report
with. Figures and tables use short names. R1-1.5B to R1-32B are the DeepSeek-R1-Distill-Qwen bases, OMN-1.5B is OpenMath-Nemotron-1.5B, and Q3-1.7B and Q3-4B are the Qwen3 bases. JustRL-D and JustRL-N are JustRL-DeepSeek and JustRL-Nemotron. Skywork-7B, Skywork-Math and Skywork-32B are Skywork-OR1-7B, Skywork-OR1-Math-7B and Skywork-OR1-32B. AceReason and AceReason-14B are AceReason-Nemotron-7B and AceReason-Nemotron-14B. The other recipes go by their names without the size.}
\label{tab:recipes}
\end{table}
\begin{table}[!htbp]
\ContinuedFloat
\centering\scriptsize
\setlength{\tabcolsep}{2.5pt}\renewcommand{\arraystretch}{1.0}
\begin{tabular}{>{\raggedright\arraybackslash}p{2.3cm}>{\raggedright\arraybackslash}p{1.6cm}>{\raggedright\arraybackslash}p{2.8cm}>{\raggedright\arraybackslash}p{2.5cm}>{\raggedright\arraybackslash}p{2.3cm}>{\raggedright\arraybackslash}p{1.5cm}>{\raggedright\arraybackslash}p{0.9cm}>{\raggedright\arraybackslash}p{1.6cm}}
\toprule
model & base & algorithm, scale & length shaping & RL data & training cap & rollout $T$ & authors' evaluation \\
\midrule
JustRL-DeepSeek-1.5B & R1-Distill-1.5B & GRPO without KL or entropy bonus, clip-higher; $4{,}380$ steps & none; fixed $16$k context & $17$k math (DAPO-Math-17k) & $15$k & $1.0$ & $32$k, $T{=}0.7$ \\
DeepMath-1.5B & R1-Distill-1.5B & GRPO with a KL loss, clip-higher, dynamic sampling; $1{,}800$ steps & overlong penalty, $0$ to $-1$ over $20$k to $24$k; truncated $-2$ & $103$k math (DeepMath-103K) & $24$k & $0.6$ & $32$k, $T{=}0.6$ \\
RLinf-math-1.5B & R1-Distill-1.5B & GRPO without KL; steps n.d. & none & $106$k math, with $10$k logic puzzles & $27.6$k & $1.0$ & $32$k, $T{=}0.6$ \\
AReaL-1.5B-Preview-Stage-3 & R1-Distill-1.5B, through two released earlier stages & critic-free PPO with a small KL; about $435$ steps over three stages & none; cap raised $8$k$\to16$k$\to24$k & $40$k math (DeepScaleR set) & $8$k to $24$k & $0.6$ & $32$k, $T{=}0.6$ \\
JustRL-Nemotron-1.5B & OpenMath-Nemotron-1.5B & as JustRL-DeepSeek; $3{,}440$ steps & none; fixed $16$k context & $17$k math (DAPO-Math-17k) & $15$k & $1.0$ & $32$k, $T{=}0.7$ \\
QuestA-Nemotron-1.5B & OpenMath-Nemotron-1.5B & GRPO without KL, dynamic sampling; a partial solution in every prompt ($50\%$, then $25\%$ of it); $2{,}000$ steps & none & $9$k hard math (OpenR1-Math) & $24$k & $1.0$ & $32$k, $T{=}0.7$ \\
KnowRL-Nemotron-1.5B & OpenMath-Nemotron-1.5B & GRPO without KL, clip-higher, dynamic sampling; knowledge-point hints in $78\%$ of prompts; $2{,}960$ steps & none & $9$k math (QuestA's problems; a third multiple choice) & $24$k (paper), $32$k (card) & $1.0$ & $32$k, $T{=}0.9$ \\
DeepMath-Omn-1.5B & OpenMath-Nemotron-1.5B & as DeepMath-1.5B; $700$ steps & overlong penalty, $0$ to $-1$ over $20$k to $24$k; truncated $-2$ & $103$k math (DeepMath-103K) & $24$k & $0.6$ & $32$k, $T{=}0.6$ \\
e3-1.7B & Qwen3-1.7B & GRPO with a KL loss, clip-higher; steps n.d. & none; cap raised $8$k$\to16$k with harder data & $15$k math (DeepScaleR set) & $8$k to $16$k & $0.6$ & $32$k, $T{=}0.6$ \\
Polaris-1.7B-Preview & Qwen3-1.7B & as Polaris-4B; about $500$ steps & none; cap raised $40$k$\to48$k & $53$k math, with $4.6$k logic puzzles & $40$k to $48$k & $1.5$ to $1.55$ & $90$k, $T{=}1.4$ \\
\midrule
AceReason-Nemotron-14B & R1-Distill-14B & on-policy GRPO without KL or entropy loss; math, then code, then math & none; cap raised $8$k$\to24$k on math, $32$k on code and the last math stage & $49$k math, $8.5$k code & $8$k to $32$k & n.d. & $32$k, $T{=}0.6$ \\
Skywork-OR1-32B & R1-Distill-32B & GRPO without KL, adaptive entropy, rejection sampling; released at step $1{,}000$ of $1{,}130$ & none; cap raised $16$k$\to24$k & Skywork set, as Skywork-OR1-7B & $16$k to $24$k & $1.0$ & $32$k, $T{=}1.0$ \\
\bottomrule
\end{tabular}
\caption{(continued) The ten added recipes, registered before their outputs were generated. Below the rule are the $14$B and $32$B recipes, read in addition and not registered. QuestA and KnowRL put guidance
in the training prompt. QuestA's guidance is a prefix of a reference solution. KnowRL's is knowledge points. The hints are absent from this paper's prompts. QuestA's authors also evaluate without the hints. KnowRL's authors report both. Columns as above.}
\end{table}

\newcommand{\pairtwo}[2]{\begin{tabular}[c]{@{}c@{}}#1\\[-1.5pt]#2\end{tabular}}

\begin{table}[!htbp]
\centering\scriptsize\setlength{\tabcolsep}{4pt}
\begin{tabular}{llccccccc}
\toprule
recipe & cap & \finB{}, base$\to$RL & $F$ & $P$ & matched $P$ & reach & change [interval] & $P$, MATH-500 \\
\midrule
Open-RS3 & 4k & $0.21\to0.21$ & 0.002 & 0.001 & 0.003 & 0 & 0.003 [$-$0.007, 0.014] & $-$0.002 \\
ProRL v2 & 8k & $0.45\to0.99$ & \textbf{0.203} & \textbf{$-$0.026} & 0.023 & 0 & 0.178 [0.146, 0.211] & \textbf{$-$0.025} \\
JustRL-D & 16k & $0.82\to0.98$ & \textbf{0.063} & \textbf{0.032} & \textbf{0.048} & $-$0.003 & 0.092 [0.068, 0.118] & 0.008 \\
e3 & 16k & $0.84\to0.92$ & \textbf{0.043} & $-$0.002 & 0.009 & \textbf{$-$0.007} & 0.034 [0.013, 0.056] & \textbf{$-$0.018} \\
JustRL-N & 16k & $0.93\to0.95$ & \textbf{0.013} & \textbf{0.024} & \textbf{0.032} & 0.003 & 0.039 [0.018, 0.060] & \textbf{0.014} \\
DeepScaleR & 24k & $0.94\to0.98$ & \textbf{0.020} & \textbf{0.027} & \textbf{0.039}$^{\ast}$ & 0.004 & 0.051 [0.031, 0.071] & 0.009 \\
DeepMath & 24k & $0.94\to0.99$ & \textbf{0.025} & \textbf{0.052} & \textbf{0.071}$^{\ast}$ & 0.007 & 0.084 [0.063, 0.105] & 0.011 \\
AReaL & 24k & $0.94\to0.96$ & \textbf{0.012} & 0.008 & \textbf{0.017}$^{\ast}$ & $-$0.001 & 0.019 [0.003, 0.036] & \textbf{0.012} \\
RLinf & 28k & $0.95\to0.99$ & \textbf{0.017} & \textbf{0.036} & \textbf{0.039}$^{\ast}$ & 0 & 0.054 [0.030, 0.078] & \textbf{0.025} \\
STILL-3 & 28k & $0.95\to0.97$ & \textbf{0.010} & \textbf{0.016} & \textbf{0.020}$^{\ast}$ & 0.005 & 0.030 [0.014, 0.047] & 0.004 \\
QuestA & 24k & $0.98\to0.81$ & \textbf{$-$0.036} & \textbf{0.051} & \textbf{0.045}$^{\ast}$ & \textbf{0.006} & 0.021 [0, 0.043] & \textbf{0.025} \\
KnowRL & 24k & $0.98\to0.98$ & \textbf{0.006} & \textbf{0.037} & \textbf{0.033}$^{\ast}$ & 0 & 0.043 [0.022, 0.065] & \textbf{0.024} \\
DeepMath-Omn & 24k & $0.98\to0.97$ & 0.002 & \textbf{0.018} & 0.017$^{\ast}$ & $-$0.001 & 0.020 [0.002, 0.038] & \textbf{0.011} \\
Skywork-32B & 24k & $0.98\to0.97$ & \textbf{$-$0.009} & \textbf{0.047} & \textbf{0.037} & 0.013 & 0.052 [0.027, 0.078] & \textbf{0.010} \\
Skywork-7B & 32k & $0.99\to0.96$ & \textbf{$-$0.008} & \textbf{0.050} & \textbf{0.052}$^{\ast}$ & 0.009 & 0.051 [0.029, 0.074] & \textbf{0.025} \\
Skywork-Math & 32k & $0.99\to0.98$ & 0.001 & \textbf{0.036} & \textbf{0.027}$^{\ast}$ & 0.013 & 0.050 [0.023, 0.077] & \textbf{0.023} \\
AceReason & 32k & $0.99\to1.00$ & \textbf{0.008} & \textbf{0.022} & 0.019$^{\ast}$ & $-$0.002 & 0.028 [0.005, 0.050] & \textbf{0.016} \\
AceReason-14B & 32k & $0.99\to1.00$ & \textbf{0.006} & \textbf{0.045} & \textbf{0.042}$^{\ast}$ & 0.010 & 0.061 [0.036, 0.086] & \textbf{0.018} \\
Polaris-1.7B & 48k & $0.99\to0.90$ & \textbf{$-$0.027} & \textbf{0.023} & 0.015 & 0.003 & 0 [$-$0.022, 0.022] & -- \\
Polaris-4B & 52k & $1.00\to0.96$ & \textbf{$-$0.016} & \textbf{0.021} & 0.012 & 0.005 & 0.009 [$-$0.010, 0.029] & -- \\
\bottomrule
\end{tabular}
\caption{What RL buys at each recipe's training cap, on physics: one row per recipe, the cells behind Figure~\ref{fig:overview} and the cap rows of Figure~\ref{fig:recipes}. $F$, $P$ and net reach split the change in full-credit pass@1, so the change can differ from a mean-credit score by up to $0.03$. \textbf{Bold}: the $95\%$ paired bootstrap interval over problems excludes zero. $F$ and $P$ are computed outside the unsolved set $U$, so $F$'s sign can differ from the printed change in \finB{}. Matched $P$ is reliability on problems both models finish seven times in eight. $^{\ast}$: computed on the second generation of Table~\ref{tab:regrade2}, which agrees within $0.007$ on every term. Open-RS3's matched $P$ rests on ten problems. The Polaris pairs were not read on MATH-500 at their caps.  The cells at the next budget above the cap, MATH-500 in full and every interval are in the released tables.}
\label{tab:capcells}
\end{table}

\subsection{How the traces were read}
\label{sec:traces}
\textbf{Protocol.} We read up to twelve problems per recipe that the RL model solves and its base, despite finishing, does not. Two language-model readers (Claude Opus 5.5) see both derivations with model and grade labels, so the reading is not blind. The readers mark whether the RL derivation states a principle, method or reading the base's never states. A third reader settles disagreements. Of the $195$ wins, $42$ are grading or answer-form artifacts (full table in the released repository).

\textbf{Blind reading.} A blind, shuffled re-read of $143$ pairs asks whether each side states a step the other never does. Weighted to the $195$ wins, it finds a new step on the RL side in $46\%$ and on the base side in $50\%$, assuming the $60$ unmarked pairs read stand for the other $112$. A failed derivation lacks steps as often as the winning one, so the labelled count measures attention as much as content.

\textbf{A human check of the new steps.} One author, blind to side, checked the claimed new steps in $50$ of the $143$ pairs, drawn at random with a fixed seed. Of $45$ decided claims ($12$ unsure), $19$ hold. There is no detectable difference between sides: $10$ of $26$ for RL ($0.38$ [$0.22$, $0.57$]), $9$ of $19$ for the base ($0.47$ [$0.27$, $0.68$]). The readers overcount on both sides, so the counts above are upper bounds. The symmetry the reuse finding rests on survives. The check tests only claims made, so it gives no $\kappa$.

\textbf{Repertoire check.} A step missing from one base sample may still be in the model, so checkers searched four other base samples on the same problem. Of the $83$ steps flagged in the unblinded reading ($37$ from the first eight recipes, $46$ from the ten added), $77$ appear in one, $37$ used correctly. Six appear in none. Two of the $77$ were found only weakly and no finds were re-examined, so the count may lean toward finding them. Of the $10$ RL-side steps the human check upheld, the base's samples hold $9$. Of the $80$ blind RL-side steps, the search finds $58$, $32$ of them used correctly in a base sample that then abandons them. The search misses $13$, and $9$ were not searched because their base samples were not stored. The reverse search gave each blind base-side step on a physics problem, with the same prompt, to a reader with four samples of the RL model from the $16$ regenerated pairs. The reverse search finds $52$ of the $62$, $15$ of them used correctly, and misses $10$. The regenerated pairs hold no math or e3 texts, so $9$ base-side steps were not searched. On physics the forward search finds $49$ of $60$. Beyond the verdicts, the reverse readers wrote short reports, which call five of the $52$ finds borderline. Counting the five borderline finds as misses leaves $47$ of $62$. The forward readers returned verdicts only, so the forward search has no count of borderline finds. The human check took one author one sitting of about half an hour. The same author later checked the $15$ misses the human check had not decided, blind to side and with the same two questions. Over both checks, the author upholds $2$ of the $11$ RL-side steps the forward search misses on physics, rejects $6$ and is unsure of $3$. Of the $10$ base-side steps the reverse search misses, the author upholds $4$, rejects $2$ and is unsure of $4$. The forward search's two math misses are both rejected. Counting both checks, the other model's samples hold $9$ of the $11$ upheld RL-side steps and $6$ of the $10$ upheld base-side steps on physics.

\textbf{Reliability problems.} The checks above cover lone wins. A second check covers problems where the gain is reliability: the base finishes all eight samples at $16$k and gets one to four right, and the RL model gets more right. There are $249$ such pair--problems in $15$ of the $16$ regenerated pairs. We drew $40$ problems, two or three per pair, with a fixed seed. One blind reader per problem compared an RL derivation graded right with a base derivation graded wrong, shuffled among the reverse comparisons, and named the decisive step. A fresh reader then searched the base's right and wrong samples for that step. Twelve comparisons had no step to search: the values match, and only form, units, a missing box or grading differs. Of the $28$ searched, $2$ claims were judged not decisive. Of the other $26$ steps, $25$ are used correctly in a base sample graded right, $1$ is stated only in a base sample graded wrong, and none is absent. The base's wrong samples state the step on all $26$ problems: $17$ without carrying it through, $9$ using it correctly and failing elsewhere. In the reverse direction, $35$ problems have an RL sample graded wrong that finished. Of these, $11$ differ only in answer form or grading, and $24$ were searched. In $5$ of the $24$, the search reader judged the claimed step not decisive. Of the $19$ decisive base steps, $17$ are used correctly in an RL sample graded right, $2$ are stated only, and none is absent. A step that decides correctness should appear in most right samples, so the wrong samples are the informative part. Each comparison had one reader. Readers did not see the gold answers. The labels are the matcher's, not the judge's.

\textbf{Two more strata, at each recipe's cap.} The same check, forward only, ran at the caps of the $13$ recipes in the second generation that are not cap-bound. We drew $40$ of the $270$ pair--problems where the base finishes all eight samples and is right in five to seven, and the RL model is right more often. There, $12$ comparisons differ only in answer form or grading, and $2$ claims were judged not decisive. All $26$ decisive RL steps are used correctly in a base sample graded right. The base's other wrong samples state $8$ of them without carrying them through, use $5$ and lack $2$. On $11$ problems the base has no other wrong sample. We also took all $32$ distinct problems among the $72$ pair--problems where the base is never right but leaves a sample unfinished, and the RL model is right at least once. There, $5$ comparisons differ only in form or grading, and $2$ claims were judged not decisive. The compared base derivation finished in all $32$. Of the $25$ decisive RL steps, the base's own samples state $17$ without carrying them through and use $8$ correctly while failing elsewhere. None is absent.

\textbf{On the cap-bound recipes' $Z$, the base's samples write $31$ of the $32$ deciding steps.} We drew $40$ of the $68$ distinct problems among the $84$ $Z$ pair--problems of ProRL v2, Open-RS3 and JustRL-DeepSeek, by the same rules. Each base sample was given whole, up to $64$k tokens, with the point where the cap falls marked. The compared base derivation was cut at the cap on $15$ of the $40$. Of the $40$ comparisons, $6$ differ only in form or grading. In $1$ more, the base derivation has RL's physics but is cut before it answers. One claim was judged not decisive. Within the cap, the base's own samples use $24$ of the $32$ decisive RL steps correctly and state $7$ without carrying them through. The one step absent, even at $64$k tokens, weights two hydrogen lines by $2J{+}1$ to get their intensity ratio. The RL derivation there takes the ratio from a rule it does not derive. Given $64$k tokens, a base sample graded right uses the step on $15$ of the $32$, and on all $8$ of Open-RS3's. Each comparison and each search in these checks had one reader, and no reader saw the gold answers.

\subsection{Judge regrade}
\label{sec:regrade}
\textbf{The split survives a judge regrade.} Under judge protocol v2, a pinned Qwen3-32B regraded part by part every stored physics sample the frozen matcher scored below one. The judge is stricter than an earlier, lenient language-model audit of answers the matcher rejected: on $2{,}800$ answers both read, the audit alone calls $440$ right and the judge alone $55$. The answer-form share of the reliability term lies between two estimates: the judge lowers the term by $8\%$ (\S\ref{sec:discussion}), and $12$ of $40$ sampled reliability wins differ only in answer form or grading (\S\ref{sec:traces}). Net reach moves by less than $0.01$ in every cell (full table in the released repository). For the $32$B pair it falls at the cap from $0.013$ to $0.004$, because $29$ of the $68$ problems the base left unsolved in eight samples under the matcher are solved under the judge.

\textbf{The same holds on a regenerated read of sixteen of the pairs.} The judge moves $F$ by at most $0.021$, $P$ by at most $0.011$ and net reach by at most $0.008$ (Table~\ref{tab:regrade2}). No reliability term changes sign. AceReason-Nemotron-14B's net reach at its cap falls from $0.016$ to $0.010$. For the $7$B recipes the judge leaves net reach in place or raises it. For these, the answer-form reading of \S\ref{sec:reach} rests on the robust cells of Appendix~\ref{sec:boundary}, not on the regrade.

\textbf{The second generation agrees with the main read within $0.007$.} Split like Table~\ref{tab:capcells}, it moves $F$ by at most $0.004$, $P$ by at most $0.006$ and net reach by at most $0.007$, each inside the bootstrap half-width. Its reliability term is lower by $0.002$ on average. Only Open-RS3's reliability term changes sign, from $+0.001$ in the main read to $-0.002$.

\begin{table}[!htbp]
\centering\scriptsize\setlength{\tabcolsep}{5pt}
\begin{tabular}{llccc}
\toprule
recipe & cap & $F$ & $P$ & net reach \\
\midrule
ProRL v2 & 8k & $+0.219\to+0.240$ & $-0.021\to-0.026$ & $+0.001\to-0.001$ \\
JustRL-D & 16k & $+0.058\to+0.063$ & $+0.032\to+0.028$ & $-0.004\to-0.007$ \\
DeepScaleR & 24k & $+0.015\to+0.017$ & $+0.040\to+0.039$ & $-0.002\to+0.000$ \\
DeepMath & 24k & $+0.019\to+0.021$ & $+0.070\to+0.069$ & $+0.004\to+0.006$ \\
RLinf & 28k & $+0.012\to+0.013$ & $+0.039\to+0.038$ & $+0.000\to+0.002$ \\
STILL-3 & 28k & $+0.006\to+0.006$ & $+0.023\to+0.021$ & $+0.000\to+0.000$ \\
AReaL & 24k & $+0.009\to+0.010$ & $+0.012\to+0.012$ & $-0.005\to-0.002$ \\
Open-RS3 & 4k & $+0.002\to+0.002$ & $-0.003\to-0.007$ & $+0.000\to+0.000$ \\
JustRL-N & 16k & $+0.006\to+0.007$ & $+0.032\to+0.025$ & $+0.002\to+0.000$ \\
QuestA & 24k & $-0.047\to-0.051$ & $+0.062\to+0.057$ & $+0.004\to+0.004$ \\
KnowRL & 24k & $-0.000\to-0.000$ & $+0.046\to+0.036$ & $+0.000\to-0.001$ \\
DeepMath-Omn & 24k & $+0.000\to+0.000$ & $+0.016\to+0.005$ & $-0.008\to-0.010$ \\
Skywork-7B & 32k & $-0.014\to-0.016$ & $+0.060\to+0.065$ & $+0.007\to+0.011$ \\
AceReason-14B & 32k & $+0.001\to+0.001$ & $+0.066\to+0.058$ & $+0.016\to+0.010$ \\
Skywork-Math & 32k & $-0.002\to-0.003$ & $+0.048\to+0.049$ & $+0.014\to+0.013$ \\
AceReason & 32k & $+0.007\to+0.007$ & $+0.030\to+0.039$ & $+0.002\to+0.010$ \\
\bottomrule
\end{tabular}
\caption{A second generation of $16$ pairs on physics at each recipe's cap, each cell matcher $\to$ judge (judge protocol v2). Each model wrote one $64$k generation at temperature $0.6$, read at the cap by truncation, eight samples per problem on the $300$ physics problems. Its columns use partial credit and split the whole change without taking reach out first, unlike Table~\ref{tab:capcells}, which takes reach and its mirror out first and adds the mirror back to $P$. The $16$k rows and the reliabilities behind each term are in the released tables.}
\label{tab:regrade2}
\end{table}

\subsection{Authors' temperature}
\label{sec:temp}

\textbf{For the eight pairs re-read, the split does not depend on the sampling temperature.} Nine recipes report at $0.7$ to $1.4$ where we read at $0.6$. Each of the eight pairs re-read at the authors' settings splits as at $0.6$ (full table in the released repository). At its cap, Skywork-OR1-32B's net reach falls from $0.013$ [$-0.001$, $+0.029$] to $+0.002$ [$-0.010$, $+0.015$], as in the regrade. The reach it showed at $0.6$ was sampling noise and answer form. Polaris-1.7B at $1.4$ finishes $97\%$ of its derivations at the cap rather than $90\%$. Read at $1.4$ against its base at $0.6$, Polaris-1.7B is $0.019$ [$0.001$, $0.037$] ahead. Across the other six recipes, at-cap reliability runs from $+0.012$ to $+0.065$, against $+0.019$ to $+0.051$ in the re-read's own samples at $0.6$. The lowest, Polaris-4B's $+0.012$ [$-0.005$, $+0.028$], has an interval that includes zero. KnowRL, which reports $0.9$, is the one recipe above $0.6$ not re-read.

\subsection{Robust reach by recipe}
\label{sec:boundary}
\textbf{Ordering.} Rows are ordered by robust reach: wins in at least four of eight samples on problems the base leaves unsolved in eight.

\textbf{A tenth to a third of the gain at the cap falls on problems the base never gets right but leaves some samples unfinished.} Call these problems $Z$: outside $U$, with no correct base sample and at least one unfinished. Eq.~\eqref{eq:reach} books RL's gain on $Z$ to $F$ and $P$. We read $Z$ on the $16$ regenerated pairs, one $64$k run per model read at each budget. Their $F$, $P$ and reach at the cap are each within $0.007$ of Table~\ref{tab:capcells}. For the $13$ of them that are not cap-bound, RL's pass@1 on $Z$ is $10\%$ of their summed gain. Counting $Z$ as unsolved would move their net reach by at most $0.008$ and their robust reach by at most two problems. For the three cap-bound recipes, $Z$ carries $34\%$ of the summed gain. ProRL v2 alone carries $0.066$ of its $0.177$ there. Counted as unsolved, $Z$ would raise ProRL v2's net reach from $0.001$ to $0.067$ and JustRL-DeepSeek's from $-0.004$ to $0.013$. Open-RS3's would stay at about zero. ProRL v2's robust reach on this read would rise from $2$ to $21$ problems. Split within $Z$ by Eqs.~\eqref{eq:F}--\eqref{eq:P}, RL's gain there is $13\%$ finishing and $87\%$ reliability for the $13$ recipes, and $60\%$ and $40\%$ for the three cap-bound ones. The base is never right on $Z$, so the reverse order of the split books all of that gain to reliability, and the even split halves the finishing share. The fourth cap-bound recipe, e3, has no second-generation texts, so its $Z$ is not measured. Given $64$k tokens, the bases solve $6$ of the $72$ $Z$ problems RL gets right in the $13$ recipes, and $37$ of the $84$ in the three. On $31$ of the $84$, the base finishes all eight samples and gets every one wrong. The RL model is right in at least four of eight samples on $5$ of these $31$. Step checks on $32$ of the $Z$ problems RL gets right in the $13$ recipes and $40$ in the three cap-bound ones find the base's own samples writing RL's deciding step on $56$ of the $57$ where a step decides (\S\ref{sec:traces}).

\textbf{Most RL lone wins at $16$k are sampling.} On four bases (R1-Distill-Qwen 1.5B and 7B, OpenMath-Nemotron-1.5B and Qwen3-1.7B), with $56$ extra samples, the base solves $87$ of the $121$ physics pairs where a recipe beat a base that left the problem unsolved in eight samples. The $87$ include $12$ of the $18$ the RL model solves in at least four of eight. Only $3$ real physics wins remain among the $121$. In the crossing test, which compares RL's gain on problems the base never solves in eight samples with the base's gain on problems the RL model never solves, the net difference is $-0.013$ (full table in the released repository).

\textbf{The residue recurs across recipes on two of the three bases with three or more recipes.} The residue is the RL lone wins the base finished eight times and got wrong all eight. Shared residue problems exceed independent draws on R1-Distill-7B ($10$ against $5.1$ [$2$, $8$]) and OpenMath-Nemotron-1.5B ($8$ against $2.1$ [$0$, $5$]). On R1-Distill-1.5B they fall short ($11$ against $16.2$ [$13$, $20$]). One problem is residue for all eight recipes. Robust cells recur the same way. A problem every recipe on a base flips is a property of the base and the matcher, not of a recipe.

\textbf{Most robust cells are answer form.} Of $26$ robust cells read, about three fifths are answer form, such as $0.000225$~g against $0.225$~mg. About a quarter are near-misses. A handful are one base's systematic errors. One recurring cell is a real change in what the policy samples. No cell is a physics fact every base lacks.

\begin{table}[!htbp]
\centering\footnotesize\setlength{\tabcolsep}{6pt}
\begin{tabular}{llcccc}
\toprule
 & & \multicolumn{2}{c}{solved in 4+ of 8 samples} & \multicolumn{2}{c}{solved at least once} \\
recipe & read at & by RL, never by base & by base, never by RL & by RL & by base \\
\midrule
Skywork-32B & 24k & 11 of 111 & 1 of 100 & 18 & 14 \\
Skywork-Math & 32k & 9 of 122 & 2 of 114 & 21 & 16 \\
RLinf & 28k & 5 of 141 & 1 of 180 & 25 & 28 \\
Skywork-7B & 32k & 5 of 122 & 0 of 102 & 24 & 15 \\
AceReason & 32k & 5 of 122 & 2 of 129 & 22 & 20 \\
KnowRL & 24k & 3 of 201 & 4 of 199 & 28 & 23 \\
AceReason-14B & 32k & 3 of 126 & 0 of 118 & 22 & 12 \\
ProRL v2 & 8k & 3 of 34 & 0 of 167 & 4 & 12 \\
DeepScaleR & 24k & 2 of 129 & 0 of 176 & 21 & 24 \\
JustRL-N & 16k & 2 of 158 & 1 of 179 & 18 & 14 \\
Polaris-1.7B & 48k & 2 of 114 & 1 of 70 & 17 & 11 \\
Polaris-4B & 52k & 2 of 86 & 0 of 77 & 14 & 15 \\
QuestA & 24k & 1 of 201 & 0 of 89 & 22 & 7 \\
DeepMath & 24k & 1 of 129 & 1 of 170 & 19 & 14 \\
STILL-3 & 28k & 1 of 141 & 0 of 144 & 19 & 12 \\
AReaL & 24k & 1 of 129 & 0 of 150 & 17 & 15 \\
JustRL-D & 16k & 1 of 80 & 2 of 160 & 10 & 16 \\
DeepMath-Omn & 24k & 0 of 201 & 2 of 205 & 16 & 17 \\
e3 & 16k & 0 of 90 & 1 of 128 & 10 & 13 \\
Open-RS3 & 4k & 0 of 17 & 0 of 19 & 0 & 0 \\
\bottomrule
\end{tabular}
\caption{Robust wins by recipe, with the base's.   Each count is of problems one model solves while the other, finishing all eight samples, never does. ``11 of 111'' means the RL model solves $11$ of the $111$ problems its base leaves unsolved in eight. Physics and MATH-500 are pooled at each recipe's cap, eight samples a side. ProRL v2 is read at its $8$k cap and Open-RS3 at $4$k, just above its cap. The JustRL pairs, trained at $15$k, and e3 are read at $16$k. The Polaris pairs are read on physics only.}
\label{tab:robustreach}
\end{table}

\subsection{Reach at $k=32$}
\label{sec:reach32}
\textbf{Sampled $24$ more times, the bases recover their lone wins at least as often as the RL models do.} Each RL model wrote $24$ more samples on the physics part of $U$ (Table~\ref{tab:robustreach}), and each base on the mirror set (full table in the released repository). The RL models solve at least one on $23\%$ of these problems, the bases on $27\%$. The lone-win counts at $k=8$ are what sampling produces. The mirror is noise of the same size. For recipes trained at $16$k and below, the base reads at a larger budget, so its recovery there is an upper bound.

\subsection{A harder, unfiltered physics set}
\label{sec:hardset}

\textbf{Reach stays rare on problems the filter never touched.} The paper's $300$ physics problems were filtered to ones the controlled run's base solves sometimes, so reach there is a lower bound. On two further sets without that limit, all twenty pairs split the same way (Table~\ref{tab:hardcells}). Net reach at the cap is $-0.006$ to $+0.018$, clear of zero for seven pairs, touching zero for four and below zero for one. Bases leave two to three times more problems finished in all eight samples but never solved than on the paper's sets (Table~\ref{tab:robustreach}). Even so, the RL model solves at least four of eight on at most $10$ of them.

\textbf{The sets.} The \emph{harder set} has $471$ closed-form problems from OlympiadBench, PHYBench, PHYSICS's olympiad tier and FrontierScience. The \emph{unfiltered set} has $200$ random problems from the paper's pools, excluding its $300$. On AceReason-Nemotron-7B the judge leaves net reach unchanged.

\textbf{Far from ceiling, the split is the same.} The gain is small and mostly reliability where both models finish, and finishing for recipes trained at $16$k or below. Reach counts only finished samples, so QuestA and Polaris-1.7B, which finish less, could show less reach. At the next budget above the cap, where both finish more, their filtered-set net reach falls (released tables).

\textbf{The two knowledge-guided recipes' robust wins hold up when read.} QuestA and KnowRL have $10$ and $9$ robust wins on this set at their caps, on $13$ distinct problems. For each win, a language-model reader compared the base's eight answers and the RL model's eight with the gold. No base answer agrees with the gold, so none of the $19$ is answer form. On three wins, two of QuestA's and one of KnowRL's, the RL answers the judge credited do not agree with the gold. On three more, one credited answer is wrong, which leaves three right in eight. The other $13$ stand. Without the wrongly credited answers, net reach is about $0.016$ for QuestA and $0.014$ for KnowRL, still the largest. Other recipes' wins were not read this way.

\textbf{The filter did not hide reach.} Scored by the matcher alone, net reach at the cap is $-0.009$ to $+0.014$ on the unfiltered $200$, against $-0.007$ to $+0.013$ on the filtered $300$ (full table in the released repository). Net reach on OlympiadBench (February $2024$, before every base) is no larger than on FrontierScience (December $2025$, after every model) ($-0.013$ to $+0.027$ against $-0.005$ to $+0.035$), so there is no contamination signature. 

\textbf{With $24$ samples, about half of the lone wins are reached by the other side.} Over the $19$ pairs resampled, six of them only in part, the base solves $244$ of $448$ RL wins ($54\%$). The RL model solves $187$ of $289$ base wins ($65\%$). After both are resampled, the RL model solves $162$ of its wins again in its new samples while the base still fails them, and the base does the same on $70$ of its own, a net of $92$, about five per pair.


\begin{table}[!htbp]
\centering\footnotesize\setlength{\tabcolsep}{4pt}
\begin{tabular}{llcccccc}
\toprule
recipe & cap & \finB{}, base$\to$RL & $F$ & $P$ & net reach & robust net & \begin{tabular}[c]{@{}c@{}}RL solves 4+ of 8,\\ never by base\end{tabular} \\
\midrule
ProRL v2 & 8k & $0.32\to0.99$ & \textbf{0.137} & \textbf{$-$0.025} & $-$0.004 & $-$0.003 & 1 of 11 \\
JustRL-D & 16k & $0.75\to0.95$ & \textbf{0.044} & \textbf{0.013} & \textbf{$-$0.006} & $-$0.001 & 0 of 105 \\
DeepScaleR & 24k & $0.92\to0.97$ & \textbf{0.016} & \textbf{0.015} & $-$0.002 & 0 & 0 of 256 \\
DeepMath & 24k & $0.92\to0.98$ & \textbf{0.018} & \textbf{0.028} & 0.001 & 0 & 0 of 256 \\
RLinf & 28k & $0.94\to0.98$ & \textbf{0.015} & \textbf{0.026} & 0.005 & 0.001 & 2 of 284 \\
STILL-3 & 28k & $0.94\to0.95$ & \textbf{0.006} & 0.006 & $-$0.001 & 0 & 0 of 284 \\
AReaL & 24k & $0.92\to0.96$ & \textbf{0.011} & 0.004 & 0.001 & 0 & 0 of 256 \\
Open-RS3 & 4k & $0.12\to0.12$ & 0 & 0.001 & 0 & 0 & 0 of 1 \\
JustRL-N & 16k & $0.89\to0.95$ & \textbf{0.010} & \textbf{0.009} & \textbf{0.007} & 0.004 & 3 of 286 \\
QuestA & 24k & $0.96\to0.79$ & \textbf{$-$0.011} & \textbf{0.023} & \textbf{0.018} & 0.015 & 10 of 414 \\
KnowRL & 24k & $0.96\to0.96$ & \textbf{0.005} & \textbf{0.014} & \textbf{0.015} & 0.013 & 9 of 414 \\
DeepMath-Omn & 24k & $0.96\to0.98$ & \textbf{0.005} & 0.004 & \textbf{0.004} & 0.001 & 1 of 414 \\
e3 & 16k & $0.76\to0.87$ & \textbf{0.037} & \textbf{$-$0.018} & $-$0.005 & $-$0.001 & 2 of 181 \\
Polaris-1.7B & 48k & $0.95\to0.86$ & 0.003 & \textbf{0.019} & 0.005 & 0.001 & 2 of 378 \\
Polaris-4B & 52k & $0.99\to0.89$ & \textbf{$-$0.028} & \textbf{0.039} & \textbf{0.008} & 0.003 & 2 of 315 \\
Skywork-7B & 32k & $0.98\to0.94$ & \textbf{$-$0.008} & \textbf{0.026} & \textbf{0.009} & 0.006 & 5 of 310 \\
AceReason-14B & 32k & $0.98\to0.98$ & \textbf{0.005} & \textbf{0.019} & 0.006 & 0.001 & 4 of 287 \\
Skywork-32B & 24k & $0.97\to0.91$ & \textbf{$-$0.011} & \textbf{0.045} & \textbf{0.011} & 0.006 & 5 of 240 \\
Skywork-Math & 32k & $0.98\to0.98$ & \textbf{0.004} & \textbf{0.026} & 0.006 & 0.007 & 7 of 310 \\
AceReason & 32k & $0.98\to1.00$ & \textbf{0.009} & \textbf{0.027} & 0.006 & 0.006 & 5 of 310 \\
\bottomrule
\end{tabular}
\caption{The harder, unfiltered set at each recipe's cap. The set has $471$ harder problems and $200$ drawn from the paper's pools without the base filter. Each model writes eight samples per problem, scored by the fixed judge of Appendix~\ref{sec:judge-v2} with the deterministic checker and OlympiadBench's per-problem tolerances. Robust reach counts wins in at least four of eight, net of the mirror. \textbf{Bold}: the paired bootstrap interval over problems excludes zero. The last column counts the problems the base finishes eight times and never solves that the RL model solves in at least four of eight. Every interval is in the released tables.}
\label{tab:hardcells}
\end{table}

\subsection{A competition-math set}
\label{sec:mathset}

\textbf{Reach stays rare in domain.} Physics is out of domain for every recipe that states its mix except ProRL v2, and MATH-500 is near ceiling, so we read every pair at its cap on $276$ MathArena final-answer problems (Table~\ref{tab:mathcells}). Of these, $213$ are from 2025 and $63$ from AIME and HMMT February 2026, after every model. Net reach is $-0.002$ to $+0.021$, clear of zero for nine pairs and touching or straddling zero for eleven. The judge moves net reach by at most $+0.001$.

\textbf{Registered predictions.} Before any output existed we registered five. P1: every base finishes less than on physics, and finishing leads for the four cap-bound recipes. P2: reliability rises clear of zero, leading in most, in at least three quarters of recipes whose base finishes $90\%$ or more. P3: net reach within $[-0.02, +0.03]$, clear of zero for at most half the pairs, robust reach at most $0.02$. P4: no more reach on 2026 than on 2025. P5: AIME 2025 pass@1 within $0.08$ of the authors' comparable numbers. P2 to P5 held. P1 failed on Open-RS3, where neither model finishes more than $6\%$ at $4$k, leaving too few finished samples to split.

\textbf{Far from finishing, which term leads depends on the estimator.} For the five recipes on R1-Distill-1.5B, whose base finishes $0.74$ to $0.80$, finishing carries more of the gain than on physics. Under the matched split, reliability is larger for four of the five. The matched finishing term is a remainder that also carries accuracy changes outside the matched set.

\textbf{Reliability leads where the base finishes.} In all nine recipes whose base finishes $90\%$ or more, reliability rises clear of zero and leads under either split. On matched problems, where selection cannot act, \relB{} still rises clear of zero for every such base. For QuestA and KnowRL it rises $+0.109$ and $+0.129$, against $+0.245$ and $+0.188$ over all. Selection therefore accounts for part of each gain. A gain clear of zero remains.

\textbf{Contamination.} P4, a test of base contamination, holds for every pair (full table in the released repository). More reach on 2025, the sign of recipe contamination, is clear of zero only for STILL-3 and Skywork-OR1-7B. With only $11$ and $13$ finished-but-unsolved problems in their 2026 strata, contamination and chance cannot be told apart.

\textbf{The harness reproduces the authors' AIME 2025 numbers.} Where authors report pass@1 at our temperature and within one budget step of the cap, our read is within $0.040$ for all five models. The other gaps ($-0.160$ to $-0.011$) are budget or temperature effects.

\textbf{With $24$ samples the base solves about three fifths of the RL model's lone wins.} The base solves $99$ of $159$ ($62\%$). The RL model solves $31$ of the $60$ base wins. After both are resampled, the RL model solves $44$ of its wins again in its new samples while the base still fails them, and the base does the same on $17$ of its own, a net of $27$, about $1.4$ per pair.

\begin{table}[!htbp]
\centering\footnotesize\setlength{\tabcolsep}{4pt}
\begin{tabular}{llccccccc}
\toprule
recipe & cap & \finB{}, base$\to$RL & $F$ & $P$ & matched $P$ & net reach & robust net & \begin{tabular}[c]{@{}c@{}}RL solves 4+ of 8,\\ never by base\end{tabular} \\
\midrule
ProRL v2 & 8k & $0.20\to0.89$ & \textbf{0.227} & \textbf{$-$0.055} & 0.006 & $-$0.001 & 0 & 0 of 0 \\
JustRL-D & 16k & $0.49\to0.87$ & \textbf{0.140} & $-$0.004 & \textbf{0.046} & 0 & 0 & 0 of 8 \\
DeepScaleR & 24k & $0.74\to0.97$ & \textbf{0.070} & 0.011 & \textbf{0.043} & 0 & 0 & 1 of 31 \\
DeepMath & 24k & $0.74\to0.96$ & \textbf{0.070} & 0.011 & \textbf{0.050} & 0.005 & 0.004 & 1 of 31 \\
RLinf & 28k & $0.80\to0.96$ & \textbf{0.057} & \textbf{0.045} & \textbf{0.069} & 0.007 & 0.014 & 4 of 44 \\
STILL-3 & 28k & $0.80\to0.89$ & \textbf{0.026} & $-$0.006 & $-$0.001 & \textbf{0.008} & 0.007 & 2 of 44 \\
AReaL & 24k & $0.74\to0.81$ & \textbf{0.027} & \textbf{0.020} & \textbf{0.030} & 0.004 & 0.004 & 1 of 31 \\
Open-RS3 & 4k & $0.06\to0.06$ & 0 & 0.002 & 0 & 0 & 0 & 0 of 0 \\
JustRL-N & 16k & $0.66\to0.66$ & 0.003 & \textbf{0.071} & \textbf{0.035} & 0.003 & 0.004 & 1 of 15 \\
QuestA & 24k & $0.93\to0.63$ & \textbf{$-$0.171} & \textbf{0.189} & \textbf{0.087} & \textbf{0.009} & 0.004 & 1 of 56 \\
KnowRL & 24k & $0.93\to0.82$ & \textbf{$-$0.054} & \textbf{0.150} & \textbf{0.102} & \textbf{0.012} & 0.007 & 3 of 56 \\
DeepMath-Omn & 24k & $0.93\to0.92$ & $-$0.003 & \textbf{0.049} & \textbf{0.045} & $-$0.002 & 0 & 0 of 56 \\
e3 & 16k & $0.43\to0.62$ & \textbf{0.099} & \textbf{$-$0.049} & 0.015 & $-$0.002 & $-$0.004 & 0 of 5 \\
Polaris-1.7B & 48k & $0.98\to0.64$ & \textbf{$-$0.117} & \textbf{0.167} & \textbf{0.089} & \textbf{0.012} & 0 & 1 of 99 \\
Polaris-4B & 52k & $0.98\to0.77$ & \textbf{$-$0.105} & \textbf{0.139} & \textbf{0.050} & \textbf{0.014} & 0.004 & 1 of 64 \\
Skywork-7B & 32k & $0.91\to0.80$ & \textbf{$-$0.050} & \textbf{0.117} & \textbf{0.093} & \textbf{0.005} & 0 & 0 of 54 \\
AceReason-14B & 32k & $0.94\to0.91$ & $-$0.007 & \textbf{0.121} & \textbf{0.101} & \textbf{0.021} & 0.018 & 5 of 49 \\
Skywork-32B & 24k & $0.88\to0.83$ & $-$0.017 & \textbf{0.125} & \textbf{0.087} & \textbf{0.008} & 0.004 & 1 of 25 \\
Skywork-Math & 32k & $0.91\to0.91$ & 0.010 & \textbf{0.084} & \textbf{0.085} & \textbf{0.005} & 0 & 0 of 54 \\
AceReason & 32k & $0.91\to1.00$ & \textbf{0.050} & \textbf{0.061} & \textbf{0.074} & 0.007 & 0.007 & 2 of 54 \\
\bottomrule
\end{tabular}
\caption{The competition-math read at each recipe's cap. The set has $276$ final-answer problems from eight 2025 and 2026 contests. Each model writes eight samples per problem, scored by the fixed judge of Appendix~\ref{sec:judge-v2} after the exact-match checker. Matched $P$ is reliability on the problems both models finish at least seven times in eight, where selection into the finished set cannot act, weighted as $P$ is. Matched reliability is the per-pair version of the matched split. Robust reach counts wins in at least four of eight, net of the mirror. \textbf{Bold}: the paired bootstrap interval over problems excludes zero. The last column counts the problems the base finishes eight times and never solves that the RL model solves in at least four of eight. Every interval is in the released tables.}
\label{tab:mathcells}
\end{table}

\FloatBarrier
\section{The controlled run}
\label{app:controlled}
\textbf{Privileged information for hard groups.} Partial solutions \citep{questa,adaback}, answer-derived hints \citep{nurl}, scaffolds \citep{scafgrpo,hint}, reverse curricula \citep{r3}, hint reweighting \citep{ocgrpo}, expert traces \citep{expo,luffy} and trained hinters \citep{hill2026,tracelift2026,sage2026} report the deployed score, not whether content was learned. These methods do not control finishing. Our channels belong to these families and claim no algorithmic novelty. Process rewards \citep{mathshepherd,sgvr,srar,r1vl} need a target step, which on all-fail problems only a reference solution gives. Answer-conditioned rationales \citep{star,trice} and hindsight relabelling \citep{hir} leak the reward. We supply a setup, not an answer.

\subsection{The probe in full}
\label{sec:noinstall}\label{sec:latent}

\textbf{The run.} The \emph{control} is Qwen3-VL-8B-Thinking \citep{qwen3vl} trained with outcome RL (GSPO; \citealp{gspo,grpo}) on PhysReason \citep{physreason} and UGPhysics \citep{ugphysics,physicsr1} at a $12$k-token cap. Each \emph{arm} adds a \emph{channel} supplying the \emph{setup}, the physical modelling before any algebra: a reward, the prompt, hinted derivations, a policy-written setup, or a stronger model's questions.

\textbf{Two hard sets.} The \emph{held-out hard set} is $87$ PhysReason problems the base solves at most once in eight, never trained on, frozen before any arm under one registered protocol. A fresh base read scores $0.16$, above what selection implies, so arms are compared with the control. The \emph{training hard set} is $160$ problems the control never solved in its first thirty steps. As in \S\ref{sec:what}, the control's held-out score rises from $0.16$ to $0.26$ almost all by finishing (\finB{} $0.39\to0.69$; \relB{} flat at $0.38$).

\textbf{The missing fact is often in the model, which does not ask for it.} The right question, without the answer, nearly triples the training hard set's share solved at least in part in eight samples ($0.062\to0.175$; gold setup $0.350$). A stronger model's questions written without the reference setup give $0.200$, so the gain is not only leakage. At $64$ samples, $41\%$ is solved only with help and $34\%$ under no condition we tried. The model's own questions do no better than a generic checklist.

\textbf{No channel raises the held-out score clear of zero.} Only a pilot of the reward gave a valid read, since at full scale the policy gamed its scorer (\S\ref{sec:tail}; Appendices~\ref{sec:reward}--\ref{sec:questioner}). The setup hint leads the two control seeds by $+0.009$ and $+0.019$, below the detection floor (Table~\ref{tab:power}). Hinted derivations move it $-0.05$, beyond the seed gap. The policy-written setup moves it $-0.03$, near the detection floor. A $+2.4$-point hint-arm lead on an earlier corpus did not hold at the $16$k read. The held-out problems mostly carry figures but are given as text.

\textbf{Content is detected only where trained.} The control lifts the training hard set's pass@8 from $0.056$ to $0.344$ (\S\ref{sec:replay}). On the held-out set, drawn from another source, the control's gain is almost all finishing. On the $160$ easier text-only twins of its training problems it gains mostly by finishing. Content transfer is neither shown nor excluded (Appendix~\ref{sec:twins}). The channels' failure may therefore not be specific to them. Whether recurring techniques explain why guidance gains on competition mathematics is untested here.

\subsection{Why the knowledge is not installed}
\label{sec:tail}\label{sec:replay}

\begin{figure}[!htbp]
\centering
\begin{tikzpicture}
\begin{axis}[width=7.6cm,height=4.3cm,xmode=log,log basis x=2,xtick={8,16,32,64},xticklabels={8,16,32,64},
  xlabel={samples $k$},ylabel={pass@$k$ (\%)},ymin=0,ymax=50,legend style={font=\tiny,draw=none,fill=none,at={(0.5,-0.32)},anchor=north,legend columns=2,/tikz/every even column/.append style={column sep=0.3cm}},grid=major,tick label style={font=\scriptsize},label style={font=\scriptsize}]
\addplot+[mark=*] coordinates {(8,5.6) (16,9.4) (32,16.3) (64,24.4)};
\addplot+[mark=square*] coordinates {(8,11.3) (16,16.3) (32,21.9) (64,28.1)};
\addplot+[mark=triangle*] coordinates {(8,15.6) (16,27.5) (32,35.6) (64,43.8)};
\addplot+[mark=none,dashed,black] coordinates {(8,35.0) (64,35.0)};
\legend{$12{,}288$ cap $T{=}0.6$,$12{,}288$ cap $T{=}1.0$,$16{,}384$ cap $T{=}0.6$,gold setup in context ($k{=}8$)}
\end{axis}
\end{tikzpicture}%
\hfill%
\begin{tikzpicture}
\begin{axis}[width=7.6cm,height=4.3cm,xlabel={visit to the problem},ylabel={fresh successes per $64$},xmin=0.5,xmax=11.5,ymin=0,ymax=20,
  xtick={1,3,5,7,9,11},legend style={font=\tiny,draw=none,fill=none,at={(0.5,-0.32)},anchor=north,legend columns=2,/tikz/every even column/.append style={column sep=0.3cm}},grid=major,tick label style={font=\scriptsize},label style={font=\scriptsize}]
\addplot[thick,black,mark=*,mark size=1.2pt] coordinates {(1,7.0) (2,7.4) (3,7.9) (4,8.3) (5,8.8) (6,9.2) (7,10.6) (8,11.7) (9,12.9) (10,14.4) (11,15.4)};
\addplot[thick,orange!90!black,mark=*,mark size=1.2pt] coordinates {(1,7.3) (2,8.3) (3,7.8) (4,8.3) (5,8.6) (6,10.1) (7,10.9) (8,12.3) (9,14.0) (10,15.7) (11,17.7)};
\legend{{revisited, no replay},{revisited, with replay}}
\end{axis}
\end{tikzpicture}
\caption{Left: the panel plots pass@$k$ of the base model against sample count, on $160$ training problems never solved in the control's first thirty steps. The curves are the registered decoding, a hotter sampler and a longer cap. The plotted points are read off the first $k$ samples, not the unbiased estimates quoted in the text. The dashed line is the gold setup in context at eight samples.
Right: the panel plots fresh-rollout successes per $64$ on $60$ hard training problems solved at least once in $32$ samples at $16$k. The count is taken visit by visit during training, over the first thirty updates of each run. The curves are with and without the policy's own earlier successes replayed into the group (\S\ref{sec:replay}). Replayed rows are excluded from the count.}
\label{fig:tail}
\end{figure}
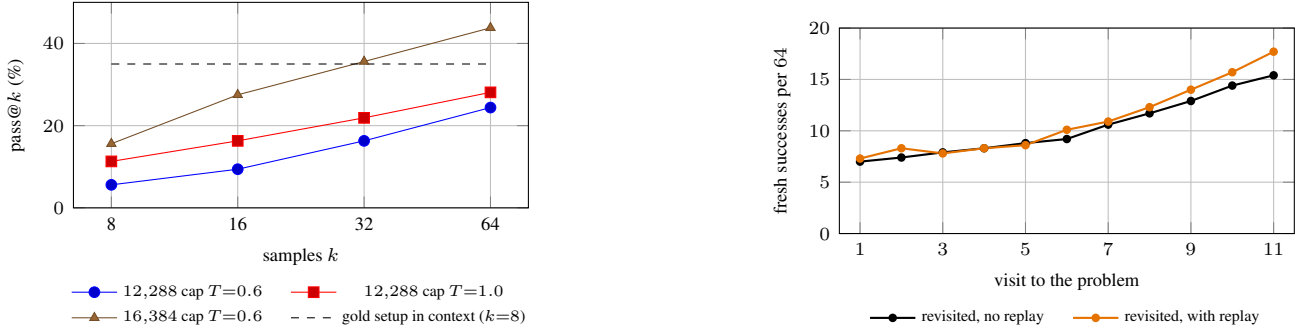

\textbf{The policy rarely samples a success.} The base solves $24\%$ of the training hard set in $64$ samples ($44\%$ at a $16$k cap), still rising (Figure~\ref{fig:tail}, left). At a few percent per sample, a group of sixteen sees a success less than half the time. At least a quarter are within reach but need a long, improbable derivation. Gains follow a $p(1-p)$ shape: $+0.16$ to $+0.18$ next visit when solved one to six times in eight, $+0.05$ and $+0.02$ at the ends.

\textbf{Finishing carries because every prompt shares it.} Finishing explains $48\%$ of within-group reward variance early and $21\%$ once saturated. Finishing's gradient adds up across the batch. Problem-specific content is not detected to carry.

\textbf{Each channel moves only what the policy already samples.} At full scale the setup reward taught unit-appending for the answer matcher, a gain that was not real. Derivations hinted with the setup lean heavily on the hint ($400$--$580$ nats; Appendix~\ref{sec:mechanism-tables}), so the bound of \citet{hill2026} gives no reason to expect transfer. A teacher's derivation lifts touched problems from $12\%$ to $59\%$, not their neighbours. Self-written setups succeed equally often (intraclass correlation $0.03$), so rewarding them changes only format. Questioner samples that ask solve no more often.

\textbf{Once a success is sampled, credit installs it, but only on that problem.} On $60$ of $353$ hard training problems, read after training at $16$k with eight samples, the solve rate goes from $0.10$ to $0.28$, or $0.31$ with replay \citep{rlep2025}. Replay only shortens the wait (Figure~\ref{fig:tail}, right). The held-out set and the $293$ unrevisited problems do not move. These problems are rarely solved, so spread to sometimes-solved problems is not excluded. Twins bear on it weakly: on the $16$ of the $60$ with twins, the revisit arm does no better than the control (Appendix~\ref{sec:twins}). The held-out bottleneck is the first success.
\subsection{Setting details}

\textbf{Setting.} The policy is Qwen3-VL-8B-Thinking \citep{qwen3vl}, given problems as text and trained with GSPO \citep{gspo} under verl \citep{verl}. Table~\ref{tab:arms} lists the arms. The reward is the fraction of gold answers matched by a boxed answer under one frozen, blind-audited matcher. The \emph{old corpus} mixes UGPhysics \citep{ugphysics,physicsr1} and PhysReason \citep{physreason} items. Every row of \emph{corpus~v2} is solvable at least sometimes by the base.

\textbf{Setup claims.} The $7{,}398$ claims are mined from answer-conditioned derivations. A blind audit found $73.8\%$ strict precision. About $1\%$ still leak a gold answer. A blind panel judged $149$ of $150$ real claims correct.

\begin{table}[!htbp]
\centering\scriptsize
\setlength{\tabcolsep}{3pt}
\begin{tabular}{>{\raggedright\arraybackslash}p{3.2cm}>{\raggedright\arraybackslash}p{1.1cm}>{\raggedright\arraybackslash}p{2.4cm}>{\raggedright\arraybackslash}p{7.0cm}>{\raggedright\arraybackslash}p{1.6cm}}
\toprule
arm & channel & corpus & what differs from its control & steps $\times$ groups \\
\midrule
outcome RL (seeds 101, 202) & --- & old & mined-milestone term, inert ($4.7\%$ mass); the comparator & $60\times96$ \\
setup-reward arm (\armF) & reward & old & judged setup-waypoint reward on all-wrong groups & $60\times8$ \\
hint arm (\armH) & prompt & old & setup block, $\Pr(\text{hint})=1-s/59$ & $60\times96$ \\
corpus~v2 control (two seeds) & --- & v2 & none; seed 2 changes rollout seed and data order only & $60\times96$ \\
hint on v2 & prompt & v2 & the hint arm's block, matched rows and draws & $60\times96$ \\
generic block & prompt & v2 & content-free block, same draws & $60\times96$ \\
guided-splice arm (\armI) & loss & old & frozen base's hinted derivations spliced into dead groups & stopped at 4 \\
teacher arm (\armJ) & loss & v2 & online hinted teacher spliced into dead student groups & $60\times96$ \\
proposer arm (\armK) & produce & v2 & self-written setup block rewarded by a frozen executor's score & stopped at 11 \\
setup-first arm v2 / v3 (\armM) & produce & v2 (+ claims) & self-written setup section judged against the reference (lenient / strict judge) & $30\times96$ each \\
deep-sampling arm (\armO) & sampling & v2 hard stratum & $24\times64$ from the control's step 60, $12$k cap (arm~O16: $16$k cap, reachable stratum) & $10$ / $15\times24$ \\
dense-revisit arm (\armD) & sampling & $60$ learnable & $24\times64$, every problem every $2.5$ steps & $8\times24$; matched rerun $30$ updates \\
replay arm (\armR) & replay & $60$ learnable & dense revisit + the policy's own successes replayed into groups & $34\times24$ \\
wide replay arm (\armRW) & replay & v2 & the control + replay ($r{=}4$ of $16$) & paused at 11 \\
questioner fine-tune (\armS) & distil & $595$ problems & supervised fine-tune on $210$ own derivations opening with the teacher's questions (+ a no-questions control) & SFT \\
16k-cap arm (\armB) & budget & v2 & response cap $16{,}384$ & $40\times96$ \\
dynamic-pool arm (\armP) & sampling & v2 + reserve & sampler retiring saturated and dead problems & $60\times96$ \\
\bottomrule
\end{tabular}
\caption{Trained arms. Groups are prompts per step. Every arm samples $16$ rollouts per prompt except the setup-reward arm ($8$). One seed per arm except the control and the comparator. The
letters are the names the appendices use. ``Learnable'' problems are the $60$ of the $353$-problem
text-only hard stratum with at least one full solve in $32$ samples at $16$k.}
\label{tab:arms}
\end{table}

\textbf{The registered endpoint.}\label{sec:protocols} The held-out hard set is PhysReason problems never used for training, mining or selection, frozen before training. The \emph{wide} set ($136$ parts) has the base right in at most one of eight samples, the \emph{strict} subset ($84$) in none. The protocol takes eight samples at temperature $0.6$, capped at $12{,}288$ tokens. The \emph{16k} variant raises the cap to $16{,}384$. The \emph{forced} variant makes every truncated derivation answer. Greedy decoding loops on this model, and $17\%$ of items flip between runs, so it is never used. Repeated reads vary by about $\pm0.01$. We do not read differences inside that band.

\textbf{Benchmarks and readout.} The benchmarks are PhysReason held-out, PhysUniBench-OE \citep{physunibench}, OlympiadBench physics \citep{olympiadbench} and PhysOlym-A \citep{physicsr1}, judged as in Appendix~\ref{sec:judge-v2}. The setup readout labels each mined claim in a rollout established, contradicted or absent (Appendix~\ref{sec:rubric}).

\subsubsection{The bottleneck is the setup, and it can be supplied}
\label{sec:bottleneck}

\textbf{Setup errors are the largest group of failures.} Of $120$ typed misses, $41.7\%$ are setup or concept and $1.7\%$ computation. Of the model's modelling decisions on $90$ held-out parts, $36.8\%$ are wrong.

\textbf{The setup can be supplied.} The share of $116$ problems with any of four samples scoring is $0.345$ unaided and $0.578$ with the mined setup. Another problem's claims add only $+0.017$ $[-0.034, +0.069]$, so $90\%$ of the lift is content. A blind grader never shown the gold separates correct from incorrect samples at AUC $0.766$.

\subsection{The control gains by finishing, and the training cap sets how much}
\label{sec:controlfinish}\label{sec:budget}

\textbf{The control gains by finishing more} (full table in the released repository). On problems finished at both the base and step $60$, \relB{} is flat (Table~\ref{tab:power}). The finished set grows to include harder problems, so new finishes are right as often as old ones.

\textbf{The training cap sets how much finishing is learned.} The $16$k-trained arm learns less finishing and gains a little matched \relB{} instead (Table~\ref{tab:power}). Read back at $12$k it falls below the control ($0.202$ against $0.242$). Caveats: one run each, the $16$k arm stopped at step $40$, and the control's step-$40$ weights were not kept, so it is read at step $40$ only from its $12$k validation pass during training.

\subsection{Statistical power}
\label{sec:power}

\textbf{No claim rests on a difference smaller than about $0.03$.} Table~\ref{tab:power}'s intervals resample problems only and leave out run-to-run variance. Adding the variance between the two control seeds (full table in the released repository) puts the minimum detectable effect near $0.03$. Against the second seed the setup-first deficits sit at the edge of the seed gap. The $16$k-cap arm's lead reads each policy at its own training cap, so it is not a same-cap contrast. Only the hint and teacher channels ran at full scale.

\begin{table}[!htbp]
\centering\scriptsize
\setlength{\tabcolsep}{3pt}
\begin{tabular}{>{\raggedright\arraybackslash}p{3.7cm}ccc}
\toprule
paired contrast (problem-level mean score, $87$ problems) & diff & $95\%$ CI & $\Pr(\le 0)$ \\
\midrule
control s60 $-$ base ($12$k) & $+0.098$ & $[+0.059, +0.142]$ & $0.000$ \\
hint on v2 s60 $-$ control s60 ($12$k) & $+0.009$ & $[-0.014, +0.032]$ & $0.232$ \\
content-free block s60 $-$ control s60 ($12$k) & $-0.003$ & $[-0.025, +0.018]$ & $0.600$ \\
teacher arm s60 $-$ control s60 ($12$k) & $-0.052$ & $[-0.076, -0.030]$ & $1.000$ \\
setup-first (lenient judge) s30 $-$ control s30 ($12$k) & $-0.029$ & $[-0.054, -0.007]$ & $0.995$ \\
setup-first (strict judge) s30 $-$ control s30 ($12$k) & $-0.036$ & $[-0.060, -0.015]$ & $1.000$ \\
$16$k-cap arm s40 ($16$k) $-$ control s40 ($12$k) & $+0.038$ & $[+0.014, +0.063]$ & $0.001$ \\
deep sampling s15 $-$ s0 ($16$k, from control s60) & $+0.003$ & $[-0.013, +0.019]$ & $0.340$ \\
replay s30 $-$ s0 ($16$k, from control s60) & $-0.007$ & $[-0.026, +0.012]$ & $0.766$ \\
\midrule
\multicolumn{4}{l}{\textit{the same arms against the second control seed}} \\
hint on v2 s60 $-$ control seed 2 s60 ($12$k) & $+0.019$ & $[-0.001, +0.039]$ & $0.033$ \\
content-free block s60 $-$ control seed 2 s60 ($12$k) & $+0.007$ & $[-0.014, +0.027]$ & $0.243$ \\
teacher arm s60 $-$ control seed 2 s60 ($12$k) & $-0.041$ & $[-0.068, -0.018]$ & $1.000$ \\
setup-first v2 s30 $-$ control seed 2 s30 ($12$k) & $-0.021$ & $[-0.042, -0.000]$ & $0.977$ \\
setup-first v3 s30 $-$ control seed 2 s30 ($12$k) & $-0.027$ & $[-0.046, -0.010]$ & $0.999$ \\
\midrule
\multicolumn{4}{l}{\textit{matched \relB{}, finished at both reads}} \\
control s60 $-$ base ($49$ problems) & $-0.000$ & $[-0.032, +0.034]$ & \\
control seed 2 s60 $-$ base ($48$ problems) & $+0.003$ & $[-0.033, +0.041]$ & \\
$16$k-cap arm s40 $-$ base ($68$ problems) & $+0.035$ & $[+0.007, +0.064]$ & \\
\bottomrule
\end{tabular}
\caption{Clustered paired bootstrap ($20{,}000$ draws) over held-out problems. Repeated reads were taken where held-out was flat. Deep sampling at $16$k has s.d.\ $0.003$ over four reads. Replay at $16$k has s.d.\ $0.008$ over seven. The matched rows read \relB{} on problems with a finished sample at both reads.}
\label{tab:power}
\end{table}

\begin{table}[!htbp]
\centering\small
\begin{tabular}{>{\raggedright\arraybackslash}p{2.6cm}>{\raggedright\arraybackslash}p{4.0cm}>{\raggedright\arraybackslash}p{4.3cm}>{\raggedright\arraybackslash}p{4.4cm}}
\toprule
channel (scale) & how the setup is given & inside training & on held-out problems \\
\midrule
reward the setup (pilot) & a judge rewards a stated setup on groups where every sample fails & no effect at the pilot's scale; at full scale the policy gamed the scorer & not tested with a valid scorer \\
\addlinespace
hint in the prompt (full) & the setup written into the prompt, faded out over training & the policy uses it: setup claims stated $55\to92\%$ & not kept once the hint is gone; a content-free block does as well \\
\addlinespace
teacher derivations (full) & correct hinted derivations spliced into groups that failed & learned on the problems it touched ($12\to59\%$ solved) & below the control ($-0.05$); no gain on similar problems \\
\addlinespace
policy writes the setup (half) & the policy writes a setup section, which is judged & the sections become correct (strict content $0.29\to0.45$) & below the control ($-0.03$); a section does not change which sample succeeds \\
\addlinespace
fine-tune on questions (supervised) & $210$ derivations that open with a stronger model's questions & the habit is learned ($99\%$ open with questions) & no better than the same derivations without the questions; held-out at the base \\
\bottomrule
\end{tabular}
\caption{Five ways of supplying the setup, and what each does. Full scale is the control's sixty steps of $96$ problems. Every comparison is against the control on the same problems or on the frozen held-out set.}
\label{tab:channels}
\end{table}

\subsection{Channel A: reward the setup}
\label{sec:reward}

\textbf{The comparator's milestone reward is inert.} The milestone reward's dense term carries $4.7\%$ of advantage mass, flips no sign and is zero $96.7\%$ of the time.

\textbf{Design and result.} All-wrong rollouts are rewarded with a frozen judge's probability that
they commit to the setup claims. At one twelfth of the control's scale every trend is flat. The
judge's AUC there is $0.655$. Signal-to-noise is $0.65$ against $2.26$.

\textbf{Caveat: at full scale the scorer was gamed.} The rerun learned to append the units the
answer matcher required, a $+99\%$ apparent gain and no real one. The rerun's absolute figures cannot be recovered. A
powered rerun with a valid scorer was never done.

\subsection{Channel B: supply the setup in the prompt}
\label{sec:used}

\textbf{The hint helps in context and is not kept.} The setup block appears with probability
$1-s/59$ at step $s$, so the final policy trains without it. Hinted rollouts lead by $+0.13$ early and do not lead late. On the $59$ problems the hint revives, a hinted correct rollout establishes
$62$ of $68$ contested claims and a late unhinted one $12$ of $57$.

\textbf{On matched data the hint shows no gain.} The hint arm is $+0.09$ [$-1.38$, $+1.75$] over its
control (full table in the released repository). The setup hint beats a content-free checklist of the same length only on PhysUniBench-OE ($+3.9$ [$+1.1$, $+6.6$]). The PhysOlym-A lead
is therefore prompt scaffolding.

\subsection{Channel C: distil the hinted derivation}
\label{sec:fix}

\textbf{Design.} When a student group is all wrong, up to $k{=}4$ correct hinted teacher rollouts
replace its last rows, rescored under the unhinted prompt. There is no KL anchor or off-context
correction.

\textbf{Channel C teaches the problems it touches.} Early-spliced problems go from $12\%$ to $59\%$ fully
solved, against $17\%$ to $31\%$ for never-spliced ones. The setup readout does not change.

\textbf{Channel C ends below the control, and untouched problems get worse.} Forced to answer on the
held-out hard set, it scores $6.2$ against the control's $8.1$, a gap of $1.9$ [$-0.2$, $+4.0$] in the control's favour. On
$371$ untouched training problems the delta is $-5.1$ [$-6.8$, $-3.5$], flat across similarity.
Outputs run about $15\%$ longer. One seed.

\subsection{Channel D: make the policy write the setup}
\label{sec:produce}

\textbf{A self-written block shows no gain.} Following TraceLift \citep{tracelift2026}, the policy's
blocks give a frozen copy $-0.008$ [$-0.054$, $+0.040$] over no block, against $+0.050$ [$-0.009$, $+0.110$] for
the gold block. A block the policy can write is one the frozen copy already assumed.

\textbf{Rewarding a written setup section changes the section, not accuracy.} A lenient judge was
gamed ($75\%$ of claims marked established against a strict grader's $34\%$). Under a strict judge
held-out hard-set accuracy is $0.202$ against the control's $0.238$ at $30$ updates. Within a
group, a more complete section does not solve more often ($+3.0$ [$-2.8$, $+5.8$]). Most wrong
derivations ($80\%$) used every reference claim.

\subsection{Channel E: distil the questioner}
\label{sec:questioner}

\textbf{Design.} We fine-tuned on $210$ correct base derivations, each opened by setup questions
Claude Sonnet~5 wrote. The control fine-tune drops the questions.

\textbf{The habit is learned and inert.} The model asks in $99.5\%$ of
samples. Asking samples solve $2.0\%$ of the time, others $2.6\%$. The gate for running RL required
training hard-set pass@8 $\ge 0.17$ and held-out at or above the base. The fine-tune fails both, so
no RL was run from it.

\subsection{Reliance, corpus and the hard sets}
\label{sec:mechanism-tables}

\textbf{Correct hinted derivations lean too heavily on the hint for the transfer bound to bite.}
The law $p \geq p_h \exp(-\rho_c)$ bounds unhinted success by hinted success and the reliance
$\rho_c$ of correct hinted derivations. Reliance is $400$--$580$ nats for the setup hint and
$530$--$600$ for spliced teacher derivations. The teacher arm's deficit carries the missing
off-context correction as a caveat.

\textbf{The two hard sets come from different sources.} The training hard set is text-only and
UGPhysics-derived ($156$ of $160$). The held-out set is PhysReason. Testing recurring content would
need a held-out split where it recurs, or text-only twins. The twins were read
(Appendix~\ref{sec:twins}). The split was not.

\textbf{More samples, a longer cap and a hotter sampler all raise pass@$k$ on the training hard set} (Figure~\ref{fig:tail}).
Each doubling of $k$ multiplies pass@$k$ by $1.67$, $1.53$ and $1.39$. The longer cap lets more
derivations finish ($0.34\to0.53$).

\subsection{Replay}
\label{sec:extra-tables}

\textbf{Replay refills failed rows with the policy's own past successes} (Table~\ref{tab:replay}).
Replay runs on $60$ hard problems, against a matched no-replay rerun.

\begin{table}[!htbp]
\centering\small
\resizebox{\columnwidth}{!}{%
\begin{tabular}{lcccccc}
\toprule
fresh-rollout successes per $64$ & visit 1 & 2 & 3 & 6 & 9 & 11 \\
\midrule
revisited, no replay ($30$ updates) & $7.0$ & $7.4$ & $7.9$ & $9.2$ & $12.9$ & $15.4$ \\
revisited, with replay ($30$ updates) & $7.3$ & $8.3$ & $7.8$ & $10.1$ & $14.0$ & $17.7$ \\
\bottomrule
\end{tabular}}
\caption{Revisiting $60$ hard problems, with and without replay. Each cell counts successes among $64$ fresh samples at one visit during training. Replayed samples are excluded.}
\label{tab:replay}
\end{table}

\textbf{New successes are mostly fresh derivations.} A late success's nearest replayed row has
median shingle Jaccard $0.039$, against $0.034$ for its nearest early success. The held-out hard
set reads $0.329$ without replay and $0.325$ with it, against the control's $0.324$.
\subsection{Text-only twins of the training hard set}
\label{sec:twins}

\textbf{Design.} Each of the $160$ training hard-set problems has a twin (same physics and method, new surface) whose gold two blind solvers reproduce. The \emph{credited} problems are the $75$ where the control's first seed earned reward in steps $31$--$60$. The committed plan compares credited and other twins' gains over the base in mean score and any-solve@8. The split below, with its matched contrast as the content-transfer test, was chosen after a first look at $113$ of the $160$ problems.

\textbf{On the credited originals, finishing's share of the gain depends on the order of the split.} Finishing and reliability both rise there (Table~\ref{tab:twins}), so the term they make together is large. Eq.~\eqref{eq:F} gives that term to finishing, which then makes up $71\%$ of the control's gain ($73\%$ for the second seed). Splitting the term evenly gives finishing $50\%$ ($51\%$). Giving it to reliability leaves finishing $29\%$ (also $29\%$).

\textbf{On the planned measure, the credited twins gain more, $26$--$62\%$ of it by finishing.} The mean-score excess is $+0.096$ [$+0.008$, $+0.182$], and $+0.115$ [$+0.029$, $+0.202$] for the second seed, whose credit does not define the groups. The any-solve@8 excess, $+0.095$ [$-0.054$, $+0.242$], includes zero. Of the pass@1 excess, $+0.093$ [$+0.005$, $+0.181$], finishing makes up $0.058$ under Eq.~\eqref{eq:F}, $0.048$ with the cross term split evenly and $0.038$ with it given to reliability. The rest is $+0.035$ [$-0.011$, $+0.078$], $+0.045$ [$-0.021$, $+0.108$] and $+0.055$ [$-0.032$, $+0.140$] under the three orders. For the second seed, finishing makes up $56$, $41$ and $26\%$ of $+0.110$, and the rest is $+0.048$ [$+0.004$, $+0.090$], $+0.065$ [$-0.001$, $+0.128$] and $+0.081$ [$-0.007$, $+0.169$]. Mean score per finished sample falls on both groups. Regrouping by the second seed's credit is a weak check, since the credited sets share $53$ of $74$ problems.

\textbf{The split of Eqs.~\eqref{eq:finrel} and~\eqref{eq:reach} finds no content gain on the twins, and cannot exclude one} (Table~\ref{tab:twins}). Where both models finish at least seven times in eight, the credited-minus-other reliability contrast is $-0.003$ [$-0.048$, $+0.037$] (second seed: $-0.008$ [$-0.043$, $+0.021$]). That interval is as wide as the originals' contrast, $+0.039$ [$+0.006$, $+0.081$], on only $5$ and $10$ problems.

\textbf{A never-credited model and an adjusted contrast leave the excess unresolved.} The questioner fine-tune (E), never trained on these problems, shows a credited-minus-other twin difference of $+0.035$ [$-0.020$, $+0.091$] from the problems alone. Finishing little more than the base, E is not a clean control. The control's difference less E's is $+0.061$ [$-0.029$, $+0.149$]. Adjusting for the base's finishing and score gives $+0.077$ [$-0.004$, $+0.160$] on the twins, against $+0.186$ [$+0.132$, $+0.237$] on the originals. Neither check is clear of zero. Both checks were added after the full read.

\textbf{The twins are easier than their originals} (Table~\ref{tab:twins}), partly perhaps from answer form, since only twins whose gold the reward can match were kept. With the base finishing under half the twins, finishing dominates (\S\ref{sec:termination}). Content needed only on the hardest problems would not show.

\textbf{No channel beats the control on the twins} (full table in the released repository). On credited twins the hint (B) is level, the revisit arm (\S\ref{sec:replay}) within noise, and the teacher's derivation (C), D and E lower. C's interval includes zero. Channels are compared on all twins, not only those of problems they touched.

\textbf{In the twins read, the base solves $16$ of the $160$ originals at least once}, against $9$ in the earlier read. At such low rates the solved set varies by read.

\begin{table}[!htbp]
\centering\footnotesize
\setlength{\tabcolsep}{3.5pt}
\begin{adjustbox}{max width=\textwidth}
\begin{tabular}{llcccccccc}
\toprule
control & problems & $n$ & pass@1 & \finB{} & \relB{} & finishing $F$ & reliability $P$ & matched $P$ ($n$) & net reach \\
 & & & base$\to$control & base$\to$control & base$\to$control & & & & \\
\midrule
seed 1 & credited originals & $75$ & $0.02\to0.21$ & $0.30\to0.75$ & $0.08\to0.28$ & 0.131 & 0.034 [0.017, 0.055] & 0.033 [0, 0.074] (5) & 0.018 [0.003, 0.037] \\
 & credited twins & $75$ & $0.28\to0.55$ & $0.37\to0.76$ & $0.77\to0.72$ & 0.288 & $-$0.020 [$-$0.045, 0.006] & 0.014 [$-$0.002, 0.032] (15) & $-$0.002 [$-$0.005, 0] \\
 & other originals & $85$ & $0.01\to0.02$ & $0.38\to0.75$ & $0.02\to0.03$ & 0.013 & 0.007 [0.001, 0.014] & $-$0.006 [$-$0.016, 0.001] (10) & $-$0.006 [$-$0.015, 0.001] \\
 & other twins & $85$ & $0.36\to0.54$ & $0.45\to0.78$ & $0.81\to0.69$ & 0.231 & $-$0.053 [$-$0.084, $-$0.016] & 0.017 [$-$0.018, 0.060] (27) & $-$0.004 [$-$0.012, 0] \\
\midrule
seed 2 & credited originals & $75$ & $0.02\to0.23$ & $0.30\to0.81$ & $0.08\to0.29$ & 0.154 & 0.036 [0.017, 0.058] & 0.023 [$-$0.004, 0.081] (5) & 0.022 [0.002, 0.055] \\
 & credited twins & $75$ & $0.28\to0.57$ & $0.37\to0.78$ & $0.77\to0.73$ & 0.304 & $-$0.019 [$-$0.046, 0.009] & 0.009 [$-$0.005, 0.026] (15) & 0 [0, 0] \\
 & other originals & $85$ & $0.01\to0.04$ & $0.38\to0.77$ & $0.02\to0.05$ & 0.024 & 0.007 [$-$0.001, 0.016] & 0 [$-$0.010, 0.010] (10) & 0.001 [$-$0.003, 0.006] \\
 & other twins & $85$ & $0.36\to0.54$ & $0.45\to0.81$ & $0.81\to0.67$ & 0.243 & $-$0.059 [$-$0.089, $-$0.024] & 0.017 [$-$0.006, 0.049] (27) & $-$0.009 [$-$0.025, 0] \\
\bottomrule
\end{tabular}
\end{adjustbox}
\caption{Text-only twins: each control seed against the base on the training hard set's credited and other problems and on their text-only twins. Each read takes eight samples at a $12{,}288$-token cap, with full credit as success. The change in pass@1 splits into finishing $F$, reliability $P$ and net reach as in Eq.~\eqref{eq:reach}. Matched $P$ is reliability on the $n$ problems both models finish at least seven times in eight, weighted as $P$ is. The groups are defined by the first seed's credit in steps $31$--$60$. Splitting the twins' change this way was chosen after a first look at $113$ of the $160$ problems.}
\label{tab:twins}
\end{table}

\subsection{Why the finishing signal transfers}
\label{sec:model}
\textbf{Credit splits into a finishing term and a content term.} Write rewards $r_j=f_jc_j$ ($f_j$: finished inside the budget; $c_j$: correct once finished). The group-relative step is proportional to $\mathrm{Cov}(r_j,\nabla\log\pi(y_j))$, which scales with $p(1-p)$. By the law of total covariance,
\[
\mathrm{Cov}(r,\nabla\log\pi)=\mathrm{Cov}(f\,\mathbb{E}[c\mid f],\nabla\log\pi)+\mathbb{E}[f\,\mathrm{Cov}(c,\nabla\log\pi\mid f)].
\]
Finishing is shared by every prompt, so its term adds across a batch and carries to held-out prompts. Content adds only where a problem recurs, so it installs success on its own problem, as replay shows.

\FloatBarrier
\section{Instruments}
\label{app:instruments}
\subsection{Setup readout rubric}
\label{sec:rubric}
\textbf{Each item is a problem, its mined setup claims and one trimmed rollout.} The grader (Claude Sonnet~5) marks each claim \textsc{established}, \textsc{contradicted} (an incompatible setup) or \textsc{absent}, judging only the setup, not the arithmetic or the answer.

\subsection{The judge protocol}
\label{sec:judge-v2}
\textbf{Protocol~v2 fixes the matching rules in code.} Graders disagree mostly on whether a bare number matches a gold with a unit. Giving units tracks training: models trained with and without the benchmark's split give units $97.2\%$ and $64.0\%$ of the time. A deterministic checker decides $43$--$55\%$ of items. A model given the same rules decides the rest.

\textbf{Differences between systems are grader-invariant; absolute scores are not.} Regrading six systems on PhysReason held-out ($n{=}526$) under v2 with a pinned Qwen3-32B is $6.7$--$8.9$ points stricter on every system. The margin between systems agrees to $0.11$ points.
\end{document}